\documentclass[10pt,twocolumn,letterpaper]{article}

\usepackage[pagenumbers]{cvpr} 

\usepackage{url}
\usepackage[utf8]{inputenc} 
\usepackage[T1]{fontenc}    
\usepackage{booktabs}       
\usepackage{amsfonts}       
\usepackage{amsmath}
\usepackage{amssymb}
\usepackage{amsthm}
\usepackage{graphicx}
\usepackage{nicefrac}       
\usepackage{microtype}      
\usepackage{dsfont}
\usepackage{multirow}
\usepackage{xcolor}
\usepackage[export]{adjustbox}
\usepackage{diagbox}
\usepackage{bm}
\usepackage{algorithm}
\usepackage{algorithmic}
\usepackage{tabularx}
\usepackage{wrapfig}
\usepackage{array, boldline, makecell}
\usepackage{color, colortbl}
\definecolor{Gray}{gray}{0.9}

\definecolor{codegreen}{rgb}{0,0.6,0}
\definecolor{codegray}{rgb}{0.5,0.5,0.5}
\definecolor{codepurple}{rgb}{0.58,0,0.82}
\definecolor{backcolour}{rgb}{0.95,0.95,0.92}
\usepackage{listings}
\lstdefinestyle{mystyle}{
    backgroundcolor=\color{backcolour},   
    commentstyle=\color{codegreen},
    keywordstyle=\color{magenta},
    numberstyle=\tiny\color{codegray},
    stringstyle=\color{codepurple},
    basicstyle=\ttfamily\footnotesize,
    breakatwhitespace=false,         
    breaklines=true,                 
    captionpos=b,                    
    keepspaces=true,
    showspaces=false,                
    showstringspaces=false,
    showtabs=false,                  
    tabsize=2
}
\usepackage[pagebackref,breaklinks,colorlinks]{hyperref}

\usepackage[capitalize]{cleveref}
\crefname{section}{Sec.}{Secs.}
\Crefname{section}{Section}{Sections}
\Crefname{table}{Table}{Tables}
\crefname{table}{Tab.}{Tabs.}

\def\cvprPaperID{5537} 
\def\confName{CVPR}
\def\confYear{2022}

\newtheorem{lemm}{Lemma}

\begin{document}

\title{Learning to Affiliate: Mutual Centralized Learning for Few-shot Classification}

\author {
    Yang Liu\textsuperscript{\rm 1,2},
    Weifeng Zhang\textsuperscript{\rm 1},
    Chao Xiang\textsuperscript{\rm 1},
    Tu Zheng\textsuperscript{\rm 1,2},
    Deng Cai\textsuperscript{\rm 1,2},
    Xiaofei He\textsuperscript{\rm 1,2} \\
    \textsuperscript{\rm 1}Zhejiang University \quad
    \textsuperscript{\rm 2}Fabu Inc., Hangzhou, China \\
    \{lyng\_95, zhangwf, chaoxiang, zhengtu\}@zju.edu.cn \quad dengcai@cad.zju.edu.cn \quad hexiaofei@fabu.ai
}

\twocolumn[{%
\renewcommand\twocolumn[1][]{#1}%
\maketitle
\begin{center}
    \newcommand{\minus}{\scalebox{0.75}[1.0]{$-$}}
    {
    \setlength{\tabcolsep}{0pt}
    \begin{tabular}{cccc}
    \centering
    \includegraphics[width=0.25\textwidth]{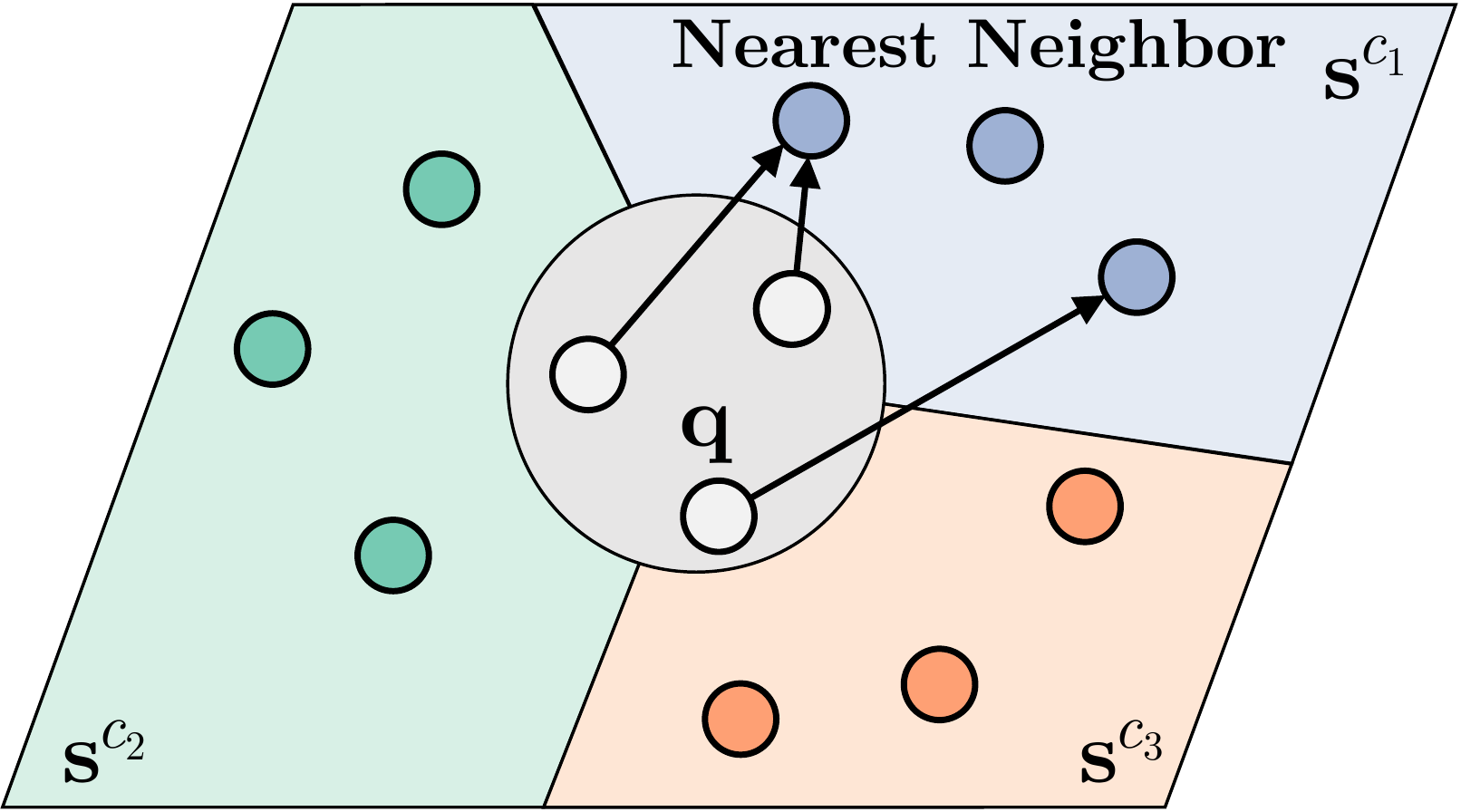} & \includegraphics[width=0.25\textwidth]{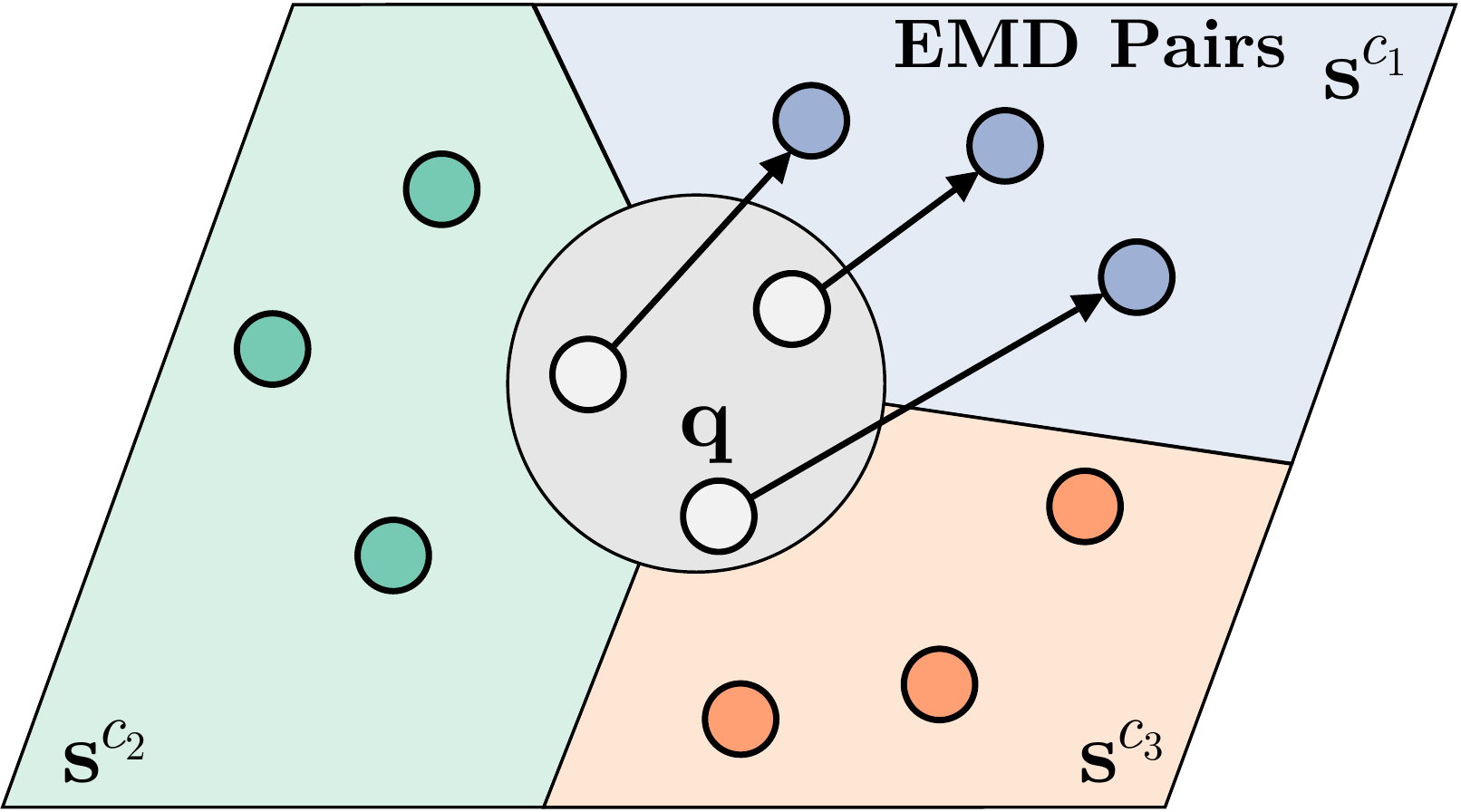} & \includegraphics[width=0.25\textwidth]{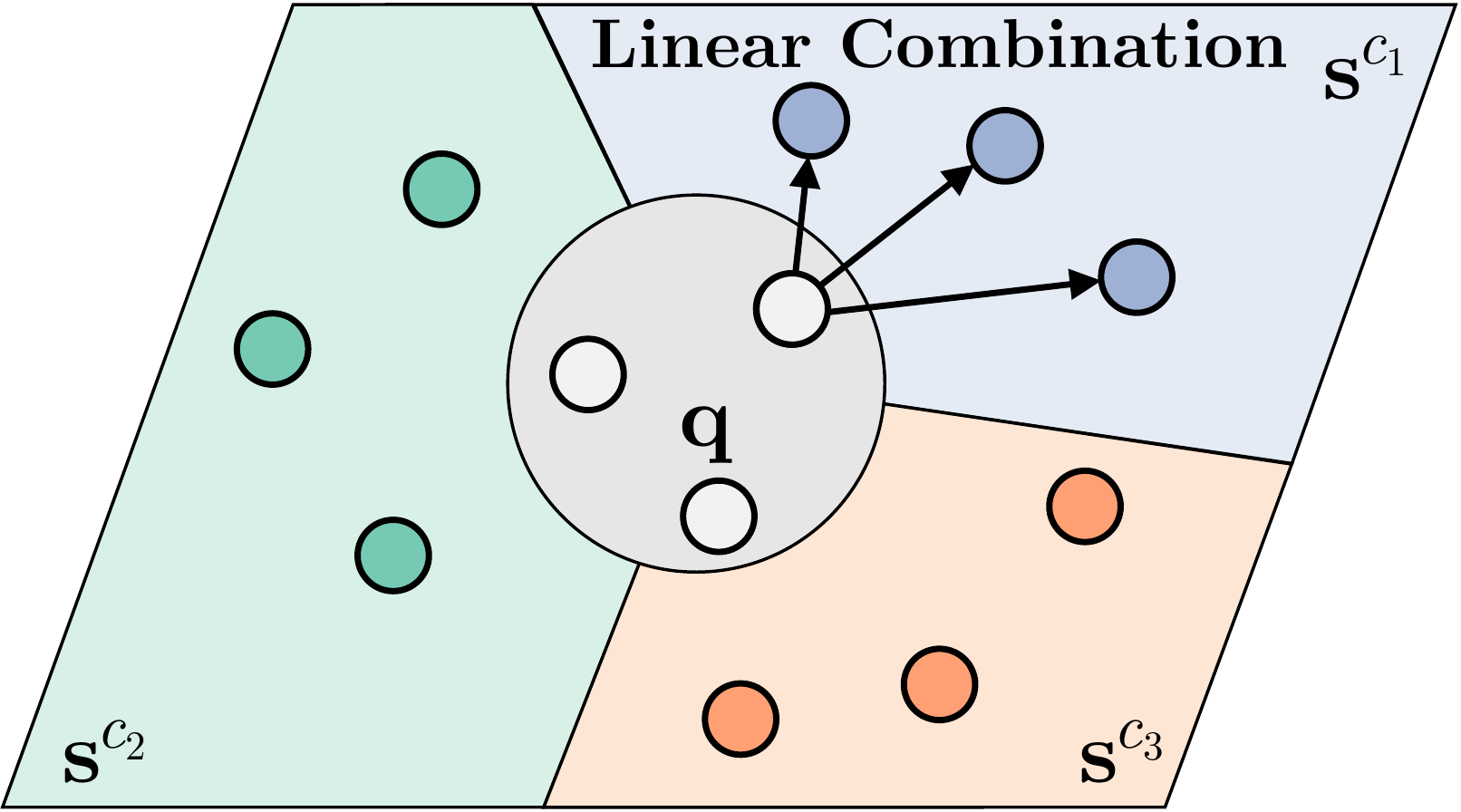} &
    \includegraphics[width=0.25\textwidth]{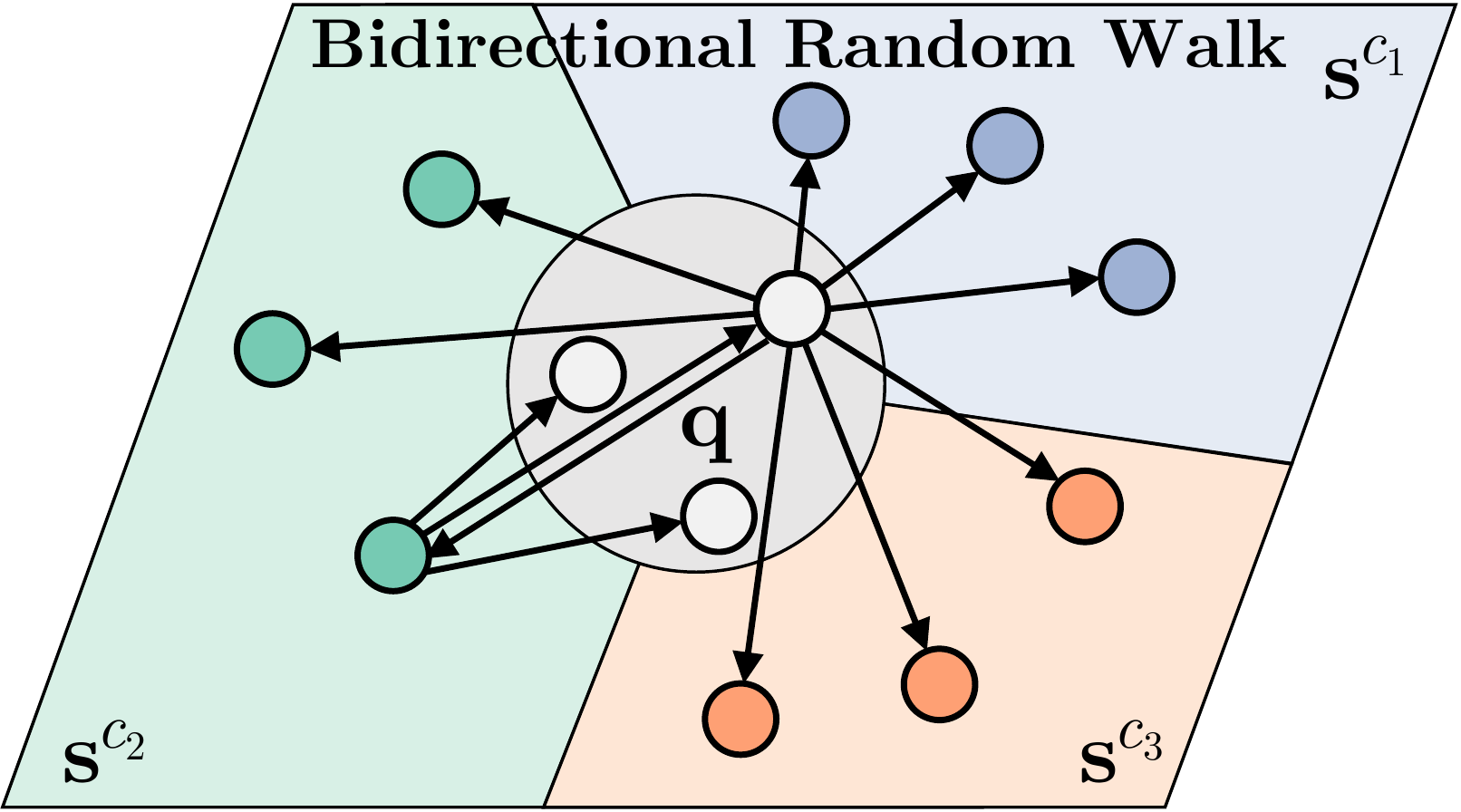} \\
    
    \multicolumn{1}{l}{\footnotesize{(a) $\mathbf{Pr}_\mathrm{DN4}(c\mid\mathbf{q},\mathbf{S}$})}&\multicolumn{1}{l}{\hspace{-5pt}\footnotesize{(b) $\mathbf{Pr}_\mathrm{DeepEMD}(c\mid\mathbf{q},\mathbf{S})$}} & \multicolumn{1}{l}{\footnotesize{(c) $\mathbf{Pr}_\mathrm{FRN}(c\mid\mathbf{q},\mathbf{S})$}}&\multicolumn{1}{l}{\footnotesize{(d) $\mathbf{Pr}_\mathrm{MCL}(c\mid\mathbf{q},\mathbf{S})$}} \\
    
    \multicolumn{1}{l}{\footnotesize{$\displaystyle{\propto \mathrm{exp}\left(\sum_{q\in \mathbf{q}}\mathrm{cos}(q, \mathrm{NN}_{\mathbf{s}^c}(q))\right)}$}}&\multicolumn{1}{l}{\hspace{-5pt}\footnotesize{$\displaystyle{\propto \mathrm{exp}\left(\minus\sum_{q\in \mathbf{q}}\|q-\mathrm{EMD}_{\mathbf{s}^c}(q)\|^2\right)}$}} & \multicolumn{1}{l}{\footnotesize{$\displaystyle{\propto \mathrm{exp}\left(\minus\sum_{q\in \mathbf{q}}\|q - \mathbf{s}^c\mathbf{w}\|^2\right)}$}} & \multicolumn{1}{l}{\hspace{-2pt}\footnotesize{ $\displaystyle{\propto\hspace{-5pt}\sum_{z\in\mathbf{q} \cup \mathbf{S}}\hspace{-4pt}\mathbb{E}\left[\sum_{t=1}^\infty\mathds{1}[X_t\in \mathbf{s}^c]\bigg| X_0=z\right]}$}}
    \end{tabular}
    }
    \captionof{figure}{
    Illustrations of four dense features based methods in a 3-way 1-shot scenario, where each colored pane indicates an image and particles within each pane indicate three feature vectors of that image. The arrow line indicates a direct connection between two feature vectors in inference. For example, (a) DN4 \cite{li2019revisiting} accumulates cosine similarities between the nearest features; (b) DeepEMD \cite{zhang2020deepemd} finds the optimal matching flow that has the minimum cost; (c) FRN \cite{wertheimer2021few} reconstructs query features with the linear combination of support features. (d) Our MCL lets query and support features bidirectionally random walk to the opposite dense features set and measures the expected number of visits in a Markov process $\{X_t\}$. Difference from them, MCL considers mutual affiliations between the dense features instead of following the unidirectional paradigm whose evident character is an accumulation over all query features followed by a softmax probability.}
    \label{fig:intro}
\end{center}%
}]

\begin{abstract}
Few-shot learning (FSL) aims to learn a classifier that can be easily adapted to accommodate new tasks, given only a few examples. To handle the limited-data in few-shot regimes, recent methods tend to collectively use a set of local features to densely represent an image instead of using a mixed global feature. They generally explore a unidirectional paradigm, \eg, finding the nearest support feature for every query feature and aggregating local matches for a joint classification. In this paper, we propose a novel Mutual Centralized Learning (MCL) to fully affiliate these two disjoint dense features sets in a bidirectional paradigm. We first associate each local feature with a particle that can bidirectionally random walk in discrete feature space. To estimate the class probability, we propose the dense features' accessibility that measures the expected number of visits to the dense features of that class in a Markov process. We relate our method to learning a centrality on an affiliation network and demonstrate its capability to be plugged in existing methods by highlighting centralized local features. Experiments show that our method achieves the new state-of-the-art.
\end{abstract}

\section{Introduction}

Few-shot classification aims to learn a classifier that can be readily adapted to novel classes given just a small number of labeled instances. To address this problem, a line of previous literature adopts metric-based methods \cite{vinyals2016matching, snell2017prototypical, sung2018learning} that learn a global image representation in an appropriate feature space and use a distance metric to predict their labels.

Recent approaches \cite{zhang2020deepemd, li2019revisiting, lifchitz2019dense} have demonstrated that the significant intra-class variations would inevitably drive the image-level embedding from the same category far apart in a given metric space under low-data regimes. In contrast, densely representative local features can provide transferrable information across categories that have shown promising performances in the few-shot scenario. Among those methods illustrated in Figure \ref{fig:intro}, DN4 \cite{li2019revisiting} finds the nearest neighbor support feature for each query feature and accumulates all the local matches in a Naive-Bayes way to represent an image-to-class similarity; DeepEMD \cite{zhang2020deepemd} uses the earth mover distance to compare the complex structured representations composed of local features. FRN \cite{wertheimer2021few} reconstructs each of query dense features with a linear combination of support dense features in a latent space and use the reconstruction distance to measure the image-to-class relevance. They all follow a unidirectional query-to-support paradigm, whose evident character is an accumulation over all query features followed by a softmax probability.

In this paper, we incorporate an extra support-to-query connection as a complement to thoroughly affiliate two disjoint sets of dense features. The potential offered by this bidirectional paradigm stems from the intuition that, except for using query features to find related support features, it is also plausible to estimate the task-relevance of query features according to the support features. Specifically, we associate dense features with particles that could bidirectionally random walk to the opposite dense features set in the discrete feature space. The prediction probability for each class is then estimated by the dense features' accessibility, \ie, the expected number of visits to the dense features of that class in a time-homogeneous Markov process. 

The contributions are as follows: (1) We propose to learn mutual affiliations between the query and support dense features instead of following the unidirectional query-to-support paradigm in FSL. (2) We introduce the dense features' accessibility to FSL and demonstrate that traditional transductive methods could be easily adapted to the inductive setting if we treat dense features from a single query image as a set of unlabeled data. (3) We propose a novel bidirectional random walk based method in FSL and draw its connection to the single-mode eigenvector centrality of an affiliation network. (4) The underlying centrality investigated in this work can be plugged in existing global feature based methods like ProtoNet and RelationNet by highlighting task-centralized local features instead of global average pooling.

\section{Related Work}

\textbf{Dense Feature based FSL.} Those methods focus on learning image-to-image similarities by encoding each input into a set of dense feature vectors (or a feature map). Among them, DC \cite{lifchitz2019dense} proposes to predict for each local features and average their probabilities. DeepEMD \cite{zhang2020deepemd} adopts the earth mover’s distance to compute an image-to-class distance. DN4 \cite{li2019revisiting} uses the top-\emph{k} nearest neighbors in a Naive-Bayes way to represent image-level similarities. ADM \cite{li2021asymmetric} and SAML \cite{hao2019collect} use extra network layers for dense classifications. FRN \cite{wertheimer2021few} reconstructs the dense query features from support features of a given class to predict their relevance. Lifchitz \etal \cite{lifchitz2021local} propagates labels from support sets to query features. MCL is also among the family of dense feature based frameworks. A major difference is that we consider the mutual affiliations between the query and support features instead of the unidirectional one in previous work.

\textbf{Graphical models in FSL.} Another branch of methods learns graphs on FSL. Among them, \cite{garcia2017few} build graph neural network (GNN) from a collection of images. \cite{ding2020graph} use the degree-centrality as part of graph features to adjust the weight of graph vertex in transductive FSL. Our proposed feature accessibility of bidirectional random walks is well-related to the eigenvector centrality. Differently, we use the single-mode centrality of bipartite data as a novel criterion for the inductive classification.

\textbf{From Transductive to Inductive.} TPN \cite{liu2019learning} first introduces the well-known label propagation \cite{zhou2003learning} to the transductive FSL that transfers information from labeled data to unlabeled one and shows substantial improvements over inductive methods. Lifchitz \etal\cite{lifchitz2021local} first demonstrates it feasible to adopt label for inductive FSL by treating the query dense features as numerous unlabeled data. Different from unidirectional label propagation, our MCL further considers the set of query and support features as bipartite data where the self-reinforcements (\ie, query-to-query and support-to-support random walks) are avoided.

\section{Method}

\subsection{Formulation of Dense Features based FSL}

In a $N$-way $K$-shot episode, the goal is to predict a class label for a single query image given $N$ support classes, each of which contains $K$ support images for reference. 

We first encode each input image into $r$ number of $d$-dimensional dense features $f_\theta(\cdot)\in \mathbb{R}^{d\times r}$. We use $\mathbf{q}=\{q_1, ..., q_r\}$ to denote dense features from a query image. Inspired by prototype learning \cite{snell2017prototypical}, we average feature vectors from the same spatial locations of $K$ different images into support dense features for each $\mathbf{s}^c=\{s_1^c, ..., s_r^c\}$ category.

We use the bold font (\eg, $\mathbf{q}$ and $\mathbf{s}$) to denote a set of feature vectors and use the normal font (\eg, $q$ and $s$) to denote a single feature vector. We use $\mathbf{S}=\bigcup_{c\in C} \mathbf{s}^c$ to represent the union set of dense features from all supporting classes. The cardinalities for the sets $\mathbf{q}$, $\mathbf{s}^c$ and $\mathbf{S}$ are $r$, $r$ and $Nr$ respectively.

The target of dense features based FSL is to find the label of a query image, given its dense query features $\mathbf{q}$ and the union set of dense support features $\mathbf{S}$ from all categories.

\subsection{Bidirectional Random Walks on Dense Features}\label{sec:2.2}

We start by using the \emph{cosine similarity} $\phi(v_1, v_2)= \langle \frac{v_1}{\|v_1\|}, \frac{v_2}{\|v_2\|} \rangle$ to measure the closeness between the two feature vectors where $\langle\cdot,\cdot\rangle$ is the Frobenius inner product. 

Given two dense features sets $\mathbf{q}$ and $\mathbf{S}$, we learn inter similarity matrix $\mathbf{\Phi}\in \mathbb{R}^{r\times Nr}$ between those two disjoint sets where the entry of $\left[\mathbf{\Phi}\right]_{qs}=\phi(q, s)$ indicates the similarity between vectors $q\in \mathbf{q}$ and $s\in \mathbf{S}$. 

We build query-to-support affiliations from every query feature $q$ to all support features $s\in \mathbf{S}$ by a random walk probability $p_{sq}=\frac{e^{\gamma\phi(s, q)}}{\sum_{s'\in \mathbf{S}}e^{\gamma\phi(s', q)}}$. The probability matrix from all query to support features can be written by
\begin{equation}\label{eq:SIM_sq}
    \mathbf{P}_\mathbf{Sq} = \mathrm{exp}(\gamma\mathbf{\Phi}^\mathrm{T}) \mathbf{D}^{-1}
\end{equation}
where $\mathrm{exp}(\cdot)$ is the element-wise exponential function, $\gamma$ is a scaling temperature for a softmax like probability and $\mathbf{D}$ is a diagonal normalization matrix with its $(j, j)$-value to be the sum of the $j$-th column of $\mathrm{exp}(\gamma\mathbf{\Phi}^\mathrm{T})$. The subscript serves both as the matrix size and as an indication for the feature vector with which it is associated. For example, $\mathbf{P}_{\mathbf{Sq}}$ is of size $|\mathbf{S}|\times |\mathbf{q}|$ and it includes random walk probabilities from query features $\mathbf{q}$ to support features $\mathbf{S}$.

Likewise, we connect every support feature $s$ to all $q\in \mathbf{q}$ by a similar probability matrix:
\begin{equation}\label{eq:SIM_qs}
    \mathbf{P}_\mathbf{qS} = \mathrm{exp}(\beta\mathbf{\Phi}) \mathbf{W}^{-1}
\end{equation}
where $\mathbf{W}$ is the analogous diagonal matrix with its $(j, j)$-value to be the sum of the $j$-th column of $\mathrm{exp}(\beta\mathbf{\Phi})$. 

It should be noted that we use two different scaling parameters since $\mathbf{\Phi}$ is a non-square matrix due to the different cardinalities of dense feature sets $\mathbf{q}$ and $\mathbf{S}$. Although the matrices $\mathrm{exp}(\gamma\mathbf{\Phi}^\mathrm{T})$ and $\mathrm{exp}(\beta\mathbf{\Phi})$ would be symmetric if we ignore their different scaling parameters, the probability matrices $\mathbf{P}_\mathbf{qS}$, $\mathbf{P}_\mathbf{Sq}$ are both column-normalized by different $\mathbf{D}$, $\mathbf{W}$ respectively and are thus directional.

To learn the mutual affiliations between the two dense feature sets, we first associate each feature vector with a particle $z$ that forms a discrete feature space $\mathbf{z} = \mathbf{q} \cup \mathbf{S}$. Next, we let particles bidirectionally random walk between particles in that space according to $\mathbf{P}_\mathbf{Sq}$ and $\mathbf{P}_\mathbf{qS}$ (\ie, particles at query features set are only allowed to move to the support features set, and vice versa). We assume it time-homogeneous in a Markov process $\{X_t\}$ and the random walk probability from $z_j$ to $z_i$ for all $z_i$, $z_j$ in $\mathbf{z}$ can be formulated by:
\begin{equation}\label{eq:3}
\resizebox{\hsize}{!}{%
$\mathbf{Pr}(X_{t+1} = z_i | X_{t}= z_j) = \left\{
\renewcommand{\arraystretch}{1.2}
\begin{array}{ll}
\displaystyle{\frac{e^{\gamma\phi(z_i, z_j)}}{\sum_{s\in \mathbf{S}} e^{\gamma\phi(s, z_j)}}} & \hspace{-5pt}z_i\in \mathbf{S}, z_j \in \mathbf{q} \\
\displaystyle{\frac{e^{\beta\phi(z_i, z_j)}}{\sum_{q\in \mathbf{q}} e^{\beta\phi(q, z_j)}}} & \hspace{-5pt}z_i\in \mathbf{q}, z_j \in \mathbf{S} \\ 0 & \hspace{-5pt}\mathrm{otherwise}.
\end{array}
\right.
$}
\end{equation}

To write it in matrix, we can obtain the Markov transition matrix $\mathbf{P}$:
\begin{equation}\label{eq:prop}
    \mathbf{P}=\begin{pmatrix}
    \mathbf{0} & \mathbf{P}_{\mathbf{S}\mathbf{q}} \\
    \mathbf{P}_{\mathbf{q}\mathbf{S}} & \mathbf{0}
    \end{pmatrix}
\end{equation}
that consists of the column-normalized $\mathbf{P}_\mathbf{Sq}$ and $\mathbf{P}_\mathbf{qS}$.

It can be proved (in supplementary) that, after infinity times of bidirectional random walks, the Markov chain with anti-diagonal $\mathbf{P}$ reaches the periodic stationary distribution:
\begin{align}\label{eq:stationary}
\begin{split}
    \lim_{t\rightarrow\infty} \mathbf{P}^{2t} &= \begin{pmatrix}
    \boldsymbol{\pi}(\mathbf{S})\bm{e}_{Nr}^\mathrm{T} & \mathbf{0} \\
     \mathbf{0} & \boldsymbol{\pi}(\mathbf{q})\bm{e}_{r}^\mathrm{T}
    \end{pmatrix} \\
    \lim_{t\rightarrow\infty} \mathbf{P}^{2t-1} &= \begin{pmatrix}
    \mathbf{0} & \boldsymbol{\pi}(\mathbf{S})\bm{e}_{r}^\mathrm{T}  \\
     \boldsymbol{\pi}(\mathbf{q})\bm{e}_{Nr}^\mathrm{T} & \mathbf{0}
    \end{pmatrix}
\end{split}
\end{align}
where $\boldsymbol{\pi}(\mathbf{S})\in \mathbb{R}^{Nr}$, $\boldsymbol{\pi}(\mathbf{q})\in \mathbb{R}^r$ are the stationary distributions with equations $\boldsymbol{\pi}(\mathbf{S}) = \mathbf{P_{Sq}}\mathbf{P_{qS}}\boldsymbol{\pi}(\mathbf{S})$ and $\boldsymbol{\pi}(\mathbf{q}) = \mathbf{P_{qS}P_{Sq}}\boldsymbol{\pi}(\mathbf{q})$ respectively. $\bm{e}_{|\cdot|}$ is a vector of ones with different length indicated by its subscript. 

The motivation of using these stationary distributions to encode mutual affiliations is straightforward: support features would be frequently visited in the long times of bidirectional random walk if they shared the same local characteristics with the query ones. In other words, the support class that owns the most mutual affiliations with querying image would be predicted due to their mutual closeness.

\subsection{Classification by Dense Features' Accessibility}

To formulate it in FSL, we first assume particles are uniformly distributed in the discrete finite space $\mathbf{z}$. Then, we use the expected number of visits from all particles $z\in \mathbf{z}$ to support features $\mathbf{s}^{c}\subset \mathbf{S}$ of class $c$ in the long times of bidirectional random walks $\{X_t\}$ to measure the amount of local characteristics that class $c$ owns for the query image:
\begin{equation}
\resizebox{\linewidth}{!}{%
$\begin{aligned}[b]\label{eq:MEL}%
&\mathbf{Pr}(\tilde{y} = c) \propto \lim_{t\rightarrow\infty}\sum_{z\in\mathbf{z}}\mathbb{E} \left[\sum_{k=1}^t\mathds{1}[X_k\in \mathbf{s}^c] \bigg| X_0=z\right]  \\ 
&= \lim_{t\rightarrow\infty} \frac{1}{t}\sum_{k=1}^{t}\left[\frac{1}{Nr+r}\sum_{s\in \mathbf{s}^c}\left(\sum_{s'\in \mathbf{S}}\left[\mathbf{P}^{2k}\right]_{ss'} + \sum_{q\in \mathbf{q}}\left[\mathbf{P}^{2k-1}\right]_{sq}\right)\right]  \\
&= \frac{1}{Nr+r} \sum_{s\in \mathbf{s}^c}\left(\sum_{s'\in \mathbf{S}}\left[\lim_{t\rightarrow \infty}\mathbf{P}^{2t}\right]_{ss'} + \sum_{q\in \mathbf{q}}\left[\lim_{t\rightarrow \infty}\mathbf{P}^{2t-1}\right]_{sq}\right)  \\ 
&=\sum_{s\in \mathbf{s}^c} \left[\boldsymbol{\pi}(\mathbf{S})\right]_s
\end{aligned}
$}
\end{equation}
where $\mathds{1}[\cdot]$ is an indicator function that equals 1 if its argument is true and zero otherwise. $[\cdot]_{ij}$ indicates the entry of the matrix from particle $j$ to particle $i$ and $[\cdot]_{i}$ indicates the entry of the vector that associated to feature $i$. 

We give the derivation when the Markov chain length $t$ is even in the first equality of Eqn.(\ref{eq:MEL}) and prove that the odd $t$ reaches the same result in the supplementary material. The second equality is from the \emph{absorbing} property of the periodic Markov chain where the power of matrix gets saturated to Eqn.(\ref{eq:stationary}) for the increasing $t$. Since $\boldsymbol{\pi}(\mathbf{S})$ is the stationary distribution of column-stochastic $\mathbf{P_{Sq}P_{qS}}$ under the probability constraint $\bm{e}^\mathrm{T}_{Nr}\boldsymbol{\pi}(\mathbf{S})=1$, $\mathbf{Pr}(\tilde{y})$ is a valid distribution for classifications.

\subsection{Reinterpretation as a Graph Centrality} \label{sec:interp}

If we interpret the random walk probability matrix $\mathbf{P}$ as an adjacency matrix $\{a_{v_1,v_2}\}$ of a directed bipartite graph $G:=(V=\{\mathbf{q}, \mathbf{S}\}, E)$, we find that the dense features' accessibility in the time-homogeneous bidirectional random walk would be equivalent to learning a \emph{single-mode eigenvector centrality} on the graph $G$. The bipartite graph is also called the \emph{affiliation network} in social network analysis \cite{bonacich1972factoring, bonacich1991simultaneous, borgatti1997network} that models two types of entities "\emph{actors}" and "\emph{society}" related by affiliation of the former in the latter. The concept of centrality in social network analysis is generally used to investigate the acquaintanceships among people that often stem from one or more shared affiliations.

To see this in graph theory, we start with a brief overview of the eigenvector centrality that reflects score $x$ of vertex $v$ for both $q\in \mathbf{q}$ and $s\in \mathbf{S}$ in the affiliation network with adjacency $\{a_{v_1,v_2}\}$:
\begin{align}\label{eq:centrality}
\begin{split}
    x_q&=\frac{1}{\lambda} \sum_{v\in \mathbb{V}(q)}a_{q, v}x_v=\frac{1}{\lambda}\sum_{v\in \mathbf{S}}a_{q, v}x_v \\      x_s&=\frac{1}{\lambda} \sum_{v\in \mathbb{V}(s)}a_{s, v}x_v=\frac{1}{\lambda}\sum_{v\in \mathbf{q}}a_{s, v}x_v
\end{split}
\end{align}
where $\mathbb{V}(\cdot)$ is a set of neighbors for the given vertex and $\lambda$ is a constant. 

With a small rearrangement, Eqn.(\ref{eq:centrality}) can be rewritten in vector notation with an eigenvector equation $\mathbf{Px} = \lambda \mathbf{x}$. The additional requirement that all the entries of the eigenvector be non-negative implies (by the Perron–Frobenius theorem \cite{perron1907theorie}) that only the greatest eigenvalue results in the desired centrality measure. For the column-stochastic adjacency matrix $\mathbf{P}$ in our method, the largest eigenvalue $\lambda$ is 1. 

Single-mode centrality \cite{borgatti1997network} is a special form of graph centrality that measures the extent to which nodes in one vertex set are relatively central only to other nodes in the same vertex set on bipartite graph. For example, the single-mode centrality for different $s$ in $\mathbf{S}$ is defined by $\hat{\mathbf{x}}_\mathbf{S}=\mathbf{x}_\mathbf{S} / \sum_{s\in \mathbf{S}}x_s$. Lemma \ref{lemm:1} (proved in supplementary material) shows that dense features' accessibility $\boldsymbol{\pi}(\mathbf{S})$ of bidirectional random walks in Eqn.(\ref{eq:MEL}) is equivalent to the single-mode eigenvector centrality of support set $\mathbf{S}$ on bipartite data.

\begin{lemm}\label{lemm:1}
Assume $G$ is the affiliation network of bipartite data $\mathbf{q}$, $\mathbf{S}$ with the adjacency matrix defined by the anti-diagonal Markov transition matrix $\mathbf{P}$ in Eqn.(\ref{eq:prop}). The single-mode eigenvector centrality $\hat{\mathbf{x}}_{\mathbf{S}}=\mathbf{x}_\mathbf{S} / \sum_{s\in \mathbf{S}}x_s$ of vertex set $\mathbf{S}$ is equivalent to the dense features' accessibility $\boldsymbol{\pi}(\mathbf{S})$ on $\mathbf{S}$ in a time-homogeneous Markov process.
\end{lemm}

Based on this reinterpretation, it is straightforward to introduce the attenuation (damping) factor $\alpha$ in Markov bidirectional random walks motivated by the Katz centrality \cite{katz1953new}, where connections made with distant neighbors are penalized by $\alpha$. The dense features' accessibility with attenuation factor $\alpha$ for few-shot classifications is defined by
\begin{align}\label{eq:katz}
\begin{split}
&\mathbf{P}\mathbf{r}_\mathrm{Katz}(\tilde{y}=c) \propto \sum_{z\in\mathbf{z}}\mathbb{E} \left[\sum_{t=1}^\infty\alpha^t\mathds{1}[X_t\in \mathbf{s}^c] \bigg| X_0=z\right]\\
&= \frac{1}{\epsilon}\sum_{t=1}^{\infty}\sum_{s\in \mathbf{s}^c}\left(\sum_{s'\in \mathbf{S}}\left[\alpha^{2t}\mathbf{P}^{2t}\right]_{ss'}+\sum_{q\in \mathbf{q}} \left[\alpha^{2t-1}\mathbf{P}^{2t-1}\right]_{sq}\right)
\end{split}
\end{align}
where $\epsilon=(Nr + r)\sum_{t=1}^{\infty}\alpha^t=(Nr+r)\alpha / (1 - \alpha)$ is a constant for a valid distribution.

Although we simply consider the single-mode centrality on $\mathbf{S}$ for an end-to-end classification purpose, it is also beneficial to learn its conjugate centrality $\boldsymbol{\pi}(\mathbf{q})$ on $\mathbf{q}$ in the affiliation network (by the analogous $\boldsymbol{\pi}(\mathbf{q})=\mathbf{x}_\mathbf{q} / \sum_{q\in\mathbf{q}}x_q$ in Lemma \ref{lemm:1}). We will show that both single-mode centralities on two dense features sets could serve as plug-and-play for finding centralized local characteristics in existing methods (hence the term Mutual Centralized Learning, MCL).

\subsection{End-to-End Training by Katz Approximation}\label{sec:MEL}

The algorithm of Mutual Centralized Learning (MCL) in Eqn.(\ref{eq:MEL}) involves the computation of Markov stationary distribution $\boldsymbol{\pi}(\mathbf{S})$ with equation $\boldsymbol{\pi}(\mathbf{S}) = \mathbf{P_{Sq}P_{qS}} \boldsymbol{\pi}(\mathbf{S})$. Theoretically, the $\boldsymbol{\pi}(\mathbf{S})$ is the eigenvector of $\mathbf{P_{Sq}P_{qS}}$ with the eigenvalue 1 under a probability constraint $\bm{e}_{Nr}^\mathrm{T}\boldsymbol{\pi}(\mathbf{S})=1$. 

The above constraints could lead to a solution of $\boldsymbol{\pi}(\mathbf{S})$ by solving the overdetermined linear system
\begin{equation}\label{eq:overdetermined}
\begin{pmatrix}
\mathbf{P_{Sq}P_{qS}} - \bm{I} \\
\bm{e}^\mathrm{T}_{Nr}
\end{pmatrix}
\boldsymbol{\pi}(\mathbf{S}) = 
\begin{pmatrix}
\mathbf{0} \\
1
\end{pmatrix}
\end{equation}
where $\bm{I}$ is an identity matrix and $\mathbf{0}$ is a vector of zeros. Although we can solve it by various methods like pseudo-inverse or QR decomposition and back substitution, it is empirically found time-consuming as these operators are either numerically unstable or not paralleled well in modern deep-learning packages.

To handle it, we present an alternative solution based on Lemma \ref{lemm:1} as follows: We first calculate the Katz centrality by its closed-form solution \cite{katz1953new}:
\begin{equation}\label{eq:katz_c}
\mathbf{x}^\mathrm{Katz} = ((\bm{I} - \alpha \mathbf{P})^{-1} - \bm{I})\bm{e}_{Nr+r}
\end{equation}
and then solve single-mode Katz centrality in Eqn.(\ref{eq:katz}) by the definition of the single-mode centrality:
\begin{equation}
    \mathbf{Pr}_\mathrm{Katz}(\tilde{y}=c) = \frac{\sum_{s\in \mathbf{s}^c}\mathbf{x}^\mathrm{Katz}_s}{\sum_{s'\in \mathbf{S}}\mathbf{x}^\mathrm{Katz}_{s'}}
    \label{eq:katz_c_2}
\end{equation}

Since Katz centrality degrades to the eigenvector centrality when $\alpha$ approaches 1 \cite{katz1953new} (indicates no attenuation), we can obtain the approximation of eigenvector centrality by a large $\alpha=0.999$:
\begin{align}\label{eq:eigen_c}
\begin{split}
    \mathbf{x}^\mathrm{Eigen} &= \lim_{\alpha\rightarrow 1}((\bm{I} - \alpha\mathbf{P})^{-1} - \bm{I})\bm{e}_{Nr+r} \\ 
    &\approx ((\bm{I} - 0.999 \mathbf{P})^{-1} - \bm{I})\bm{e}_{Nr+r}
\end{split}
\end{align}
and the analogous single-mode eigenvector centrality in Eqn.(\ref{eq:MEL}) can then be approximated by:
\begin{equation}
    \mathbf{Pr}_\mathrm{MCL}(\tilde{y}=c) = \frac{\sum_{s\in \mathbf{s}^c}\mathbf{x}^\mathrm{Eigen}_s}{\sum_{s'\in \mathbf{S}}\mathbf{x}^\mathrm{Eigen}_{s' }} \label{eq:mel_final}
\end{equation}

We use the negative log-likelihood loss to update parameters in different feature extractors and the whole picture of MCL for few-shot classifications is shown in Figure \ref{fig:arch}. We provide the detailed algorithm/pseudo-codes in the supplementary material for reference. 

\begin{figure*}[t]
    \centering
     \includegraphics[width=\textwidth]{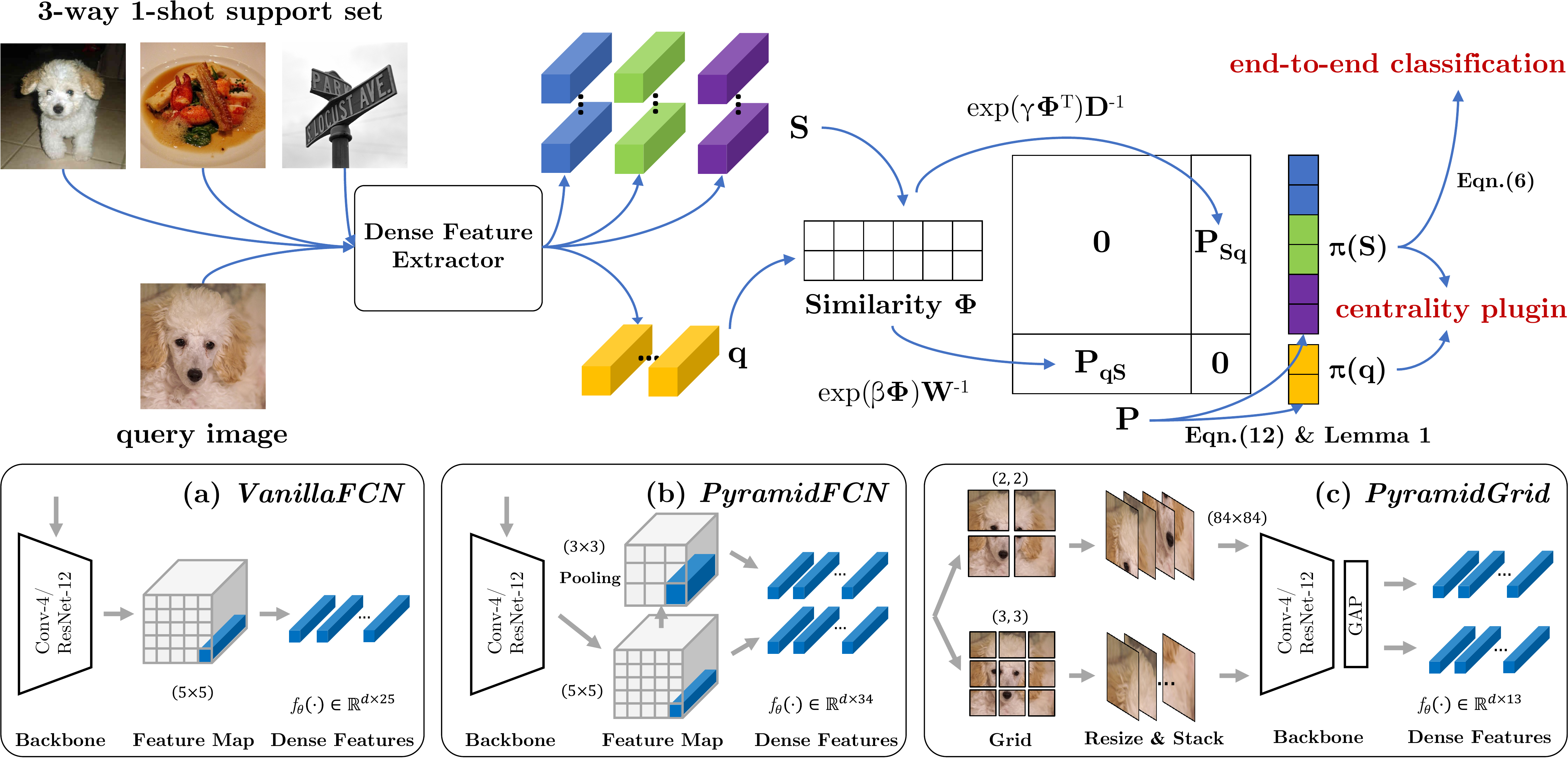}
     \caption{Our framework for 3-way 1-shot classification. Given the query and support images, we first extract their dense features $\mathbf{q}$ and $\mathbf{S}$ which are two sets of feature vectors. Then we calculate their similarity matrix $\mathbf{\Phi}$. After that, we build probability matrix $\mathbf{P}$ based on the scaled and column-normalized similarity matrix. We obtain the stationary distributions $\boldsymbol{\pi}(\mathbf{S})$ and $\boldsymbol{\pi}(\mathbf{q})$ by the Katz approximation in Eqn.(\ref{eq:eigen_c}) and the Lemma \ref{lemm:1}. Finally, we use them for both end-to-end FSL and as a centrality plugin in global feature based FSL. We explore three dense feature extractors (\ie, \textit{VanillaFCN}, \textit{PyramidFCN}, \textit{PyramidGrid}) inspired by \cite{zhang2020deepemd} to extract dense feature representations.}
     \label{fig:arch}
\end{figure*}

\subsection{Plugin for Global Feature based FSL.\label{sec:plug}}

Since the underlying centrality learned by MCL reveals the different importance of local features on an affiliation network, it is thus plausible to plug it into global feature based methods by replacing their native global average pooling (GAP) with the centrality weighted pooling as follows:

a. Extract dense features for each input image.

b. Compute the eigenvector centrality $\mathbf{x}^\mathrm{Eigen}$ by Eqn.(\ref{eq:eigen_c}).

c. Compute $\boldsymbol{\pi}(\mathbf{S})$ and $\boldsymbol{\pi}(\mathbf{q})$ according to definition of the single-mode centrality in Lemma \ref{lemm:1}.

d. Use $\boldsymbol{\pi}(\mathbf{q})$ as weights on query dense features and weighted accumulate them to a single feature vector.

e. Use the class-wise normalized centrality from $\boldsymbol{\pi}(\mathbf{S})$ to accumulate dense features for each supporting class.

Once we get the centrality weighted features for the query image and support classes, it is straightforward to perform traditional global feature based methods like before.

\section{Experiments}

We conduct experiments on two widely-used FSL datasets and three fine-grained datasets: (1) \textbf{\textit{mini}ImageNet} \cite{vinyals2016matching} contains 600 images per class over 100 classes. We follow the split used by \cite{sachin2017optimization} that takes 64, 16 and 20 classes for train/val/test respectively; (2) \textbf{\textit{tiered}ImageNet} \cite{ren2018meta} is much larger compared to \textit{mini}ImageNet with 608 classes. The 351, 97 and 160 classes are used for train/val/test respectively. (3) \textbf{CUB} \cite{welinder2010caltech} consists 11,788 images from 200 bird classes. 100/50/50 classes are used for train/val/test and each image is first cropped to a human-annotated bounding box. (4) \textbf{meta-iNat} \cite{wertheimer2019few} is a benchmark of animal species in the wild. We follow the same class split proposed by \cite{wertheimer2019few} that uses 908/227 classes for training/evaluation respectively. (5) \textbf{\textit{tiered}-meta-iNat} \cite{wertheimer2019few} is a more difficult version of meta-iNat where a large domain gap is introduced. The 354 test classes are populated by insects and arachnids, while the remaining 781 classes form the training set.

\textbf{Backbone networks.} We conduct experiments with both widely-used four layer convolutional Conv-4 \cite{vinyals2016matching} and deep ResNet-12 \cite{sun2019meta} backbones. As is commonly implemented in the state-of-the-art FSL literature, we adopt a pre-training stage for the ResNet-12 before the episode meta-training while directly meta-train from scratch for the simple Conv-4. 

\textbf{Dense feature extractor $f_\theta(\cdot)$.} We explore three dense feature extractors in our experiments as shown in Figure \ref{fig:arch}: (1) \textit{VanillaFCN} simply treats the feature map output of fully convolutional network as dense features. (2) \textit{PyramidFCN} uses an extra adaptative average pooling layer to obtain 34 dense features for each image. (3) \textit{PyramidGrid} crops the image evenly into the grid of size $2\times 2 + 3\times 3$ and encodes each grid cell to a feature vector individually. The feature vectors from all the cells constitute the set of dense features.

\begin{table*}[t]
\centering
\small
\setlength{\tabcolsep}{6pt}
\aboverulesep=0ex
\belowrulesep=0.5ex
\renewcommand\arraystretch{1}
\newcommand{\CC}{\cellcolor{Gray}}
\begin{tabular}{cccccccccc}
\toprule
\multicolumn{2}{c}{\multirow{3}{*}{\textbf{Method}}} & \multicolumn{4}{c}{\textbf{Conv-4}}                                             & \multicolumn{4}{c}{\textbf{ResNet-12}} \\
\multicolumn{2}{c}{}                        & \multicolumn{2}{c}{\textit{mini}ImageNet} & \multicolumn{2}{c}{\textit{tiered}ImageNet} & \multicolumn{2}{c}{\textit{mini}ImageNet} & \multicolumn{2}{c}{\textit{tiered}ImageNet} \\ \cmidrule(lr){3-4} \cmidrule(lr){5-6} \cmidrule(lr){7-8} \cmidrule(lr){9-10} 
\multicolumn{2}{c}{}                        & 1-shot          & 5-shot         & 1-shot           & 5-shot          & 1-shot          & 5-shot         & 1-shot           & 5-shot          \\ \midrule
\multirow{9}{*}{\textit{Global features}} & DSN \cite{simon2020adaptive} & 51.78 & 68.99 & 53.22 & 71.06 & 62.64 & 78.83 & 67.39 & 82.85 \\
 & MetaOptNet \cite{lee2019meta} & 52.87 & 68.76 & 54.71 & 71.76 & 62.64 & 78.63 & 65.99 & 81.56 \\
 & Negative Margin \cite{liu2020negative} & 52.84 & 70.41 & - & - & 63.85 & 81.57 & - & - \\
& FEAT \cite{ye2020few} & 55.15 & 71.61 & - & - & 66.78 & 82.05 & 70.80 & 84.79 \\
& Meta-Baseline \cite{chen2020new} & - & - & - & - & 63.17 & 79.26 & 68.62 & 83.29 \\
 & RelationNet$^\ddagger$ \cite{sung2018learning} & 52.12 & 66.90 & 54.33 & 69.95 & 60.97 & 75.12 & 64.71 & 78.41 \\ 
 & \CC RelationNet+MCL & \CC \textit{54.50} & \CC \textit{70.63} & \CC \textit{57.73} & \CC \textit{74.46} & \CC \textit{61.70} & \CC \textit{75.53} & \CC \textit{65.93} & \CC \textit{80.27} \\ 
 & ProtoNet$^\ddagger$ \cite{snell2017prototypical} & 52.32 & 69.74 & 53.19 & 72.28 & 62.67 & 77.88 & 68.48 & 83.46 \\
 & \CC ProtoNet+MCL & \CC \textit{54.31} & \CC \textit{69.84} & \CC \textit{56.67} & \CC \textit{74.36} & \CC \textit{64.40} & \CC \textit{78.60} & \CC \textit{70.62} & \CC \textit{83.84} \\
 \midrule
\multirow{6}{*}{\makecell{\textit{Dense features}\\(\textit{VanillaFCN})}} & DN4$^\ddagger$ \cite{li2019revisiting} & 54.66 & 71.26 & 56.86 & 72.16 & 65.35 & 81.10 & 69.60 & 83.41 \\
& DeepEMD$^\dagger$ \cite{zhang2020deepemd} & 52.15 & 65.52 & 50.89 & 66.12 & 65.91 & 82.41 & 71.16 & 83.95 \\
& FRN$^\dagger$ \cite{wertheimer2021few} & 54.87 & 71.56 & 55.54 & 74.68 & 66.45 & 82.83 & 71.16 & 86.01 \\
& Label Propagation & 52.24 & 67.68 & 54.56 & 70.08 & 65.00 & 80.07 & 71.12 & 83.89 \\
& \CC MCL (ours) & \CC 55.38 & \CC 70.02 & \CC 57.63 & \CC 74.25 & \CC 67.36 & \CC 83.63 & \CC 71.76 & \CC 86.01 \\
& \CC MCL-Katz (ours) & \CC \textbf{55.55} & \CC \textbf{71.74} & \CC \textbf{57.78} & \CC \textbf{74.77} & \CC \textbf{67.51} & \CC \textbf{83.99} & \CC \textbf{72.01} & \CC \textbf{86.02} \\
\midrule
\multirow{6}{*}{\makecell{\textit{Dense features}\\(\textit{PyramidFCN})}} & DN4$^\ddagger$ \cite{li2019revisiting} & 54.54 & 70.94 & 57.05 & 72.90 & 63.54 & 79.04 & 71.10 & 84.22 \\
& DeepEMD$^\ddagger$ \cite{zhang2020deepemd} & 50.67 & 64.94 & 51.26 & 65.64 & 66.27 & 82.41 & 70.76 & 84.20 \\
& FRN$^\ddagger$ \cite{wertheimer2021few} & 54.40 & 70.75 & 57.30 & \textbf{75.58} & 65.94 & 81.97 & 70.56 & 85.44 \\
& Label Propagation & 53.38 & 68.69 & 55.21 & 71.22 & 65.71 & 75.78 & 71.00 & 78.01 \\
& \CC MCL (ours) & \CC 55.13 & \CC 70.77 & \CC 57.93 & \CC 74.36 & \CC 67.45 & \CC 84.36 & \CC 72.01 & \CC 86.31 \\
& \CC MCL-Katz (ours) & \CC \textbf{55.77} & \CC \textbf{71.24} & \CC \textbf{58.20} & \CC 74.73 & \CC \textbf{67.85} & \CC \textbf{84.47} & \CC \textbf{72.13} & \CC \textbf{86.32} \\
\midrule
\multirow{6}{*}{\makecell{\textit{Dense features}\\(\textit{PyramidGrid})}} & DN4$^\ddagger$ \cite{li2019revisiting} & 57.17 & 70.91 & 56.71 & 70.92 & 67.86 & 80.08 & 71.29 & 82.60 \\
& DeepEMD$^\ddagger$ \cite{zhang2020deepemd} & 55.68 & 70.75 & 55.88 & 70.06 & 67.83 & 81.32 & 73.13 & 84.18 \\
& FRN$^\ddagger$ \cite{wertheimer2021few} & 55.80 & 71.52 & 55.68 & 72.87 & 67.00 & 82.20 & 71.42 & 85.58 \\
& Label Propagation & 55.12 & 68.43 & 56.05 & 72.39 & 67.18 & 81.07 & 73.18 & 85.19 \\
& \CC MCL (ours) & \CC 57.50 & \CC 73.03 & \CC 57.57 & \CC 73.81 & \CC \textbf{69.31} & \CC \textbf{85.11} & \CC \textbf{73.62} & \CC \textbf{86.29} \\
& \CC MCL-Katz (ours) & \CC \textbf{57.88} & \CC \textbf{74.03} & \CC \textbf{57.63} & \CC \textbf{73.96} & \CC 69.25 & \CC 84.71 & \CC 73.38 & \CC 86.21 \\
\bottomrule
\end{tabular}
\caption{Few-shot classification accuracy (\%) on \textit{mini}ImageNet and \textit{tiered}ImageNet. The 95\% confidence intervals are all below 0.2 for the 10,000 episodes evaluation. Results of \textit{italic} font indicates the performance of MCL as a plugin. Results of \textbf{bold} fonts are the best results for different dense feature extractors, respectively. We re-implement our plugin baseline methods (\ie, ProtoNet and RelationNet) as well as the competitive dense features based methods (\ie, DN4, DeepEMD and FRN) in our unified framework in case their performances are not available under specific settings. $\dagger$: re-implemented results on Conv-4 backbone, $\ddagger$: re-implemented results on both backbones. \label{tab:sota}}
\end{table*}

\subsection{General Few-shot Classification Results}

Table \ref{tab:sota} details the comparisons of MCL with global feature based methods \cite{simon2020adaptive,lee2019meta,ye2020few, chen2020new, liu2020negative} as well as dense feature based methods \cite{li2019revisiting, zhang2020deepemd, wertheimer2021few} on \textit{mini}-/\textit{tiered}ImageNet. 

Besides these methods, we also exploit label propagation \cite{zhou2003learning} in dense feature based inductive FSL like in \cite{lifchitz2021local} to demonstrate that traditional transductive methods could be easily adapted to inductive settings if we treat dense feature vectors as a set of unlabeled data. As shown, our MCL is highly competitive with recent state-of-the-art results. In particular, our MCL-Katz achieves \textbf{67.51}\% (1-shot) and \textbf{83.99}\% (5-shot) on \textit{mini}ImageNet with the simplest \textit{VanillaFCN}. 

\textbf{Comparisons with dense feature based methods.} The last three panes of Table \ref{tab:sota} illustrate that proposed bidirectional MCL outperforms query-to-support DN4's nearest neighboring, DeepEMD's optimal matching and FRN's latent feature reconstructions in various tasks. Although the optimal matching flow is unidirectional in DeepEMD, they adopt a cross-reference attention that encodes the bidirectional relevance to a certain extent. If we treat dense features as \textit{nodes} in the graph and use their similarities as \textit{edges}, a major difference is that they treat nodes equally to derive different weights on edges while we use the fixed edges to derive the centrality of nodes directly for classifications.

\textbf{Comparisons with global feature based methods.} The centrality investigated in this work is also capable to be plugged into global feature based methods as local features are not equally important before global average pooling. As shown in the first pane of Table \ref{tab:sota}, our proposed centrality weighted pooling could provide at most \textbf{3.4}\% and \textbf{4.5}\% performance gains for ProtoNet and RelationNet respectively. 
We show in Table \ref{tab:plug} that the improvements are also consistent if we solely apply it on the query and support dense features.

\begin{table*}[]
\centering
\small
\setlength{\tabcolsep}{5pt}
\aboverulesep=0ex
\belowrulesep=0.5ex
\renewcommand\arraystretch{1}
\begin{tabular}{ccccccccccccc}
\toprule
\multirow{3}{*}{\textbf{Method}} & \multicolumn{4}{c}{\textbf{CUB}} & \multicolumn{4}{c}{\textbf{meta-iNat}} & \multicolumn{4}{c}{\textbf{\textit{tiered}-meta-iNat}} \\
 & \multicolumn{2}{c}{Conv-4} & \multicolumn{2}{c}{ResNet-12} & \multicolumn{2}{c}{Conv-4} & \multicolumn{2}{c}{ResNet-12} & \multicolumn{2}{c}{Conv-4} & \multicolumn{2}{c}{ResNet-12} \\
\cmidrule(lr){2-3} \cmidrule(lr){4-5} \cmidrule(lr){6-7} \cmidrule(lr){8-9} \cmidrule(lr){10-11} \cmidrule(lr){12-13}
 & 1-shot & 5-shot & 1-shot & 5-shot & 1-shot & 5-shot & 1-shot & 5-shot & 1-shot & 5-shot & 1-shot & 5-shot \\
 \midrule
ProtoNet \cite{snell2017prototypical} & 71.64 & 81.50 & 82.98 & 91.38 & 53.78 & 73.80 & 76.47 & 90.20 & 35.47 & 54.85 & 47.61 & 71.06 \\
ProtoNet + MCL & \textit{76.87} & \textit{86.95} & \textit{84.59} & \textit{91.71} & \textit{57.52} & \textit{74.33} & \textit{78.75} & \textit{90.53} & \textit{39.09} & \textit{59.16} & \textit{51.22} & \textit{72.50} \\
DSN \cite{simon2020adaptive} & 66.01 & 85.41 & 80.80 & 91.19 & 58.08 & 77.38 & 78.80 & 89.77 & 36.82 & 60.11 & 46.61 & 72.79 \\
CTX \cite{doersch2020crosstransformers} & 69.64 & 87.31 & 78.47 & 90.90 & 60.03 & 78.80 & 69.04 & 88.39 & 36.83 & 60.84 & 42.91 & 69.88 \\
DN4 \cite{li2019revisiting} & 78.31 & 88.43 & 85.44 & 92.51 & 62.32 & 
79.76 & 79.58 & 91.37 & 43.82 & 64.17 & 48.99 & 72.29 \\
DeepEMD \cite{zhang2020deepemd} & 75.34 & 85.68 & 83.35 & 91.60 & 54.48 & 68.36 & 76.05 & 86.82 & 36.05 & 48.55 & 48.14 & 66.27 \\
FRN \cite{wertheimer2021few} & 73.48 & 88.43 & 83.16 & 92.59 & 62.42 & 80.45  & 73.52 & 91.83 & 43.91 & 63.36 & 48.86 & 76.62 \\
\midrule
MCL & 77.80 & 88.71 & 83.64 & 92.18 & \textbf{64.66} & \textbf{81.31} & \textbf{80.17} & 91.59 & \textbf{44.08} & \textbf{64.61} & \textbf{51.35} & \textbf{76.87} \\
MCL-Katz & \textbf{79.61} & \textbf{90.56} & \textbf{85.63} & \textbf{93.18} & 63.92 & 81.09 & 79.34 & \textbf{91.84} & 44.00 & 64.24 & 49.68 & 76.05 \\
\bottomrule
\end{tabular}
\caption{Few-shot classification (\%) results on fine-grained datasets. The 95\% confidence intervals are below 0.19 within 10,000 episodes. \label{tab:finegrained}}
\end{table*}

\begin{table}[t]
\small
\centering
\setlength{\tabcolsep}{4pt}
\renewcommand\arraystretch{1.05}
\begin{tabular}{c|c|cccc}
\multirow{2}{*}{} & \multirow{2}{*}{$f_\theta(\cdot)$}  & \multicolumn{2}{c}{\textit{mini}ImageNet} & \multicolumn{2}{c}{\textit{tiered}ImageNet} \\
 & & 1-shot      & 5-shot      & 1-shot       & 5-shot       \\ \hlineB{2}
\textit{unidirectional} & \multirow{3}{*}{\textit{VanillaFCN}} & 54.88 & 69.98   &   55.31  & 68.87 \\
MCL & & 55.38 & 70.02 & 57.63 & 74.25 \\
MCL-Katz & & \textbf{55.55} & \textbf{71.74} & \textbf{57.78} & \textbf{74.77}  \\ \hline
\textit{unidirectional}  & \multirow{3}{*}{\textit{PyramidFCN}}  & 54.92            &     69.17        &     55.69         &   69.26  \\
MCL &    & 55.13 & 70.77 & 57.93 & 74.36  \\
MCL-Katz & & \textbf{55.77} & \textbf{71.24} & \textbf{58.20} & \textbf{74.73} \\ \hline
\textit{unidirectional}     & \multirow{3}{*}{\textit{PyramidGrid}} & 56.61 & 71.40  & 55.12 & 69.19 \\
MCL & & 57.50 & 73.03 & 57.57 & 73.81  \\
MCL-Katz &  & \textbf{57.88} & \textbf{74.03} & \textbf{57.63} & \textbf{73.96}           
\end{tabular}
\vspace{-3pt}
\caption{Comparisons between the \textit{unidirectional} random walks and \textit{bidirectional} MCL on \textit{mini-}/\textit{tiered}ImageNet with Conv-4.\label{tab:uni}}
\end{table}

\subsection{Fine-grained Few-shot Classification Results}

We follow the same setting (shown in supplementary) as in FRN \cite{wertheimer2021few} to train all the models from scratch on those fine-grained datasets. We re-implement DN4 \cite{li2019revisiting}, DeepEMD \cite{zhang2020deepemd} to thoroughly compare with dense features based methods. Apart from the basic comparisons (both backbones on CUB and Conv-4 backbone on meta-iNat/\textit{tiered}-meta-iNat) in \cite{wertheimer2021few}, we conduct extra ResNet-12 experiments for all the comparing methods on meta-iNat and \textit{tiered}-meta-iNat in our unified framework for a thorough comparison.

Results in Table \ref{tab:finegrained} demonstrate that both our end-to-end MCL (\textit{VanillaFCN}) and the centrality plugin are broadly effective in various fine-grained few-shot classification tasks.

\section{Analysis}

\subsection{Ablation Study}

\textbf{Different choices of $\gamma$,$\beta$.} Since $\mathbf{\Phi}$ is non-square due to the different cardinalities of set $\mathbf{S}$ and $\mathbf{q}$, we use two different scaling parameters in Eqn.(\ref{eq:SIM_sq}) and Eqn.(\ref{eq:SIM_qs}). By the column-wise normalization, $\gamma$ and $\beta$ can be interpreted as the reciprocal of temperatures in softmax random walk probability. Large $\gamma$, $\beta$ will have a hard random walk probability that leads to a concentrated centrality in the affiliation network. However, extremely large $\gamma$, $\beta$ (\eg, one-hot probability when they approaching infinity) would inevitably bias the episodic training due to the potential gradient explosion. 

Since the cardinality of set $\mathbf{S}$ is larger than that of $\mathbf{q}$, we empirically use a larger $\gamma$ than $\beta$ as we need a harder probability in random walks from query to the large set of support features. In the experiments, we carefully select $\gamma$ and $\beta$ (generally $\gamma$\texttt{=}$20$ and $\beta$\texttt{=}$10$ for 1-shot models and $\gamma$\texttt{=}$40$ and $\beta$\texttt{=}$20$ for 5-shot models) from several combinations of parameters by their validation performance. 

\begin{table}[t]
\small
\centering
\setlength{\tabcolsep}{4pt}
\aboverulesep=0ex
\belowrulesep=0ex
\renewcommand\arraystretch{1.05}
\begin{tabular}{l|cc|cccc}
\multicolumn{1}{c|}{\multirow{2}{*}{}} & \multicolumn{2}{c|}{+MCL} & \multicolumn{2}{c}{\textit{mini}ImageNet} & \multicolumn{2}{c}{\textit{tiered}ImageNet } \\
\multicolumn{1}{c|}{}                  & $\mathbf{q}$ & $\mathbf{s}$ & 1-shot          & 5-shot         & 1-shot           & 5-shot          \\ \hlineB{2}
\multirow{4}{*}{ProtoNet \cite{snell2017prototypical}}  &    &   &  52.32     & 69.74  &        53.19          & 72.28     \\
       &  \checkmark &             &    53.74    &  69.61    & 55.36 & 73.17 \\
      &             &  \checkmark          &      53.97           &     69.74           &      56.60            &     74.18           \\
      & \checkmark   & \checkmark    &  \textbf{54.31}  & \textbf{69.84}  & \textbf{56.67} &  \textbf{74.36}               \\ \hline
\multirow{4}{*}{RelationNet \cite{sung2018learning}}  & & & 52.12 & 66.90  &    54.33   &  69.95  \\
    & \checkmark &             &      53.25           & 66.79 &     54.46 &     70.71            \\
      &      &  \checkmark           &        54.37         &  70.57 &         57.68 &         74.24        \\
      & \checkmark  & \checkmark &    \textbf{54.50} & \textbf{70.63}   & \textbf{57.73}  &    \textbf{74.46}           
\end{tabular}
\caption{Ablation of MCL as plug-and-play that individually applied on query features $\mathbf{q}$ and support features $\mathbf{s}$. The experiments are conducted with Conv-4 and \textit{VanillaFCN}\label{tab:plug}}
\end{table}

\begin{table*}[t]
\newcommand{\cellheight}{1.3cm}
\aboverulesep=0ex
\belowrulesep=0ex
\def\cellbox#1{\raisebox{-.5\height}{\includegraphics[height=\cellheight]{#1}}}
\setlength{\tabcolsep}{1pt}
\begin{tabular}{lp{0.068\textwidth}<{\centering}p{0.088\textwidth}<{\centering}cllp{0.027\textwidth{}}p{0.068\textwidth}<{\centering}p{0.088\textwidth}<{\centering}cll}
\rule{0pt}{4ex}
\raisebox{-.5\height}{\rotatebox[origin=l]{90}{\shortstack[l]{\scriptsize{ProtoNet} \\ \scriptsize{\cite{snell2017prototypical}}}}} & \multicolumn{5}{c}{\cellbox{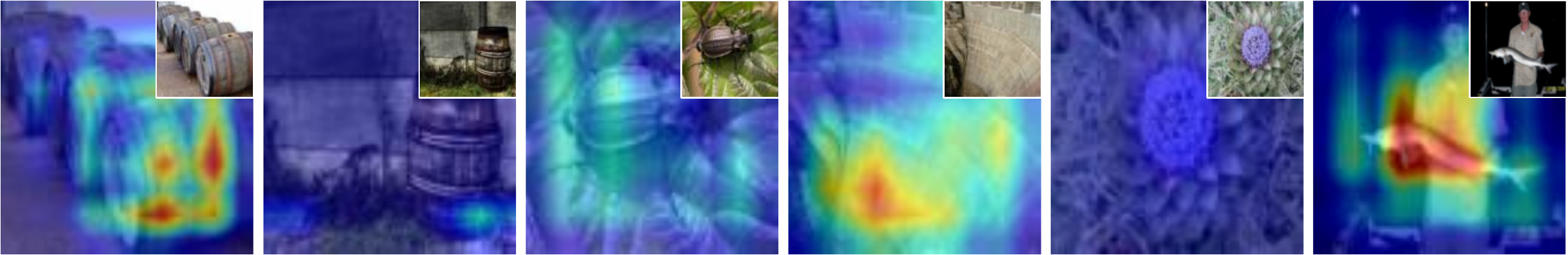}} & & \multicolumn{5}{c}{\cellbox{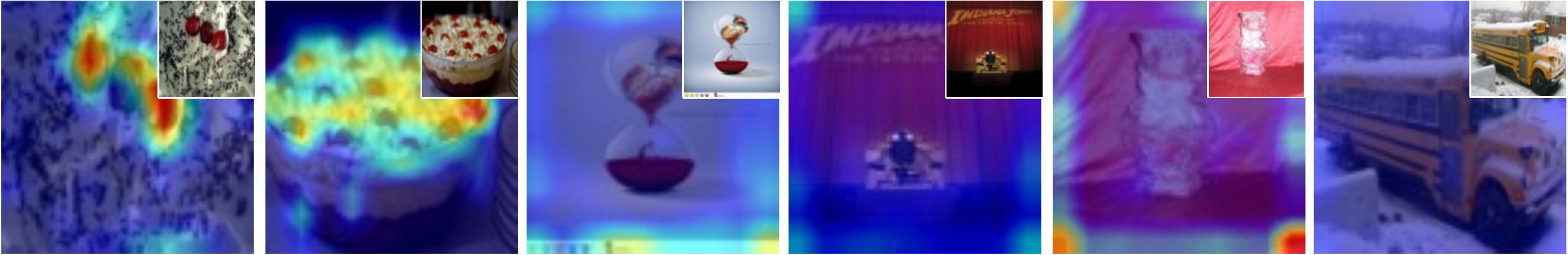}} \\ \rule{0pt}{5ex}
\raisebox{-.5\height}{\rotatebox[origin=l]{90}{\shortstack[l]{\scriptsize{ProtoNet} \\ \scriptsize{+MCL}}}} & \multicolumn{5}{c}{\cellbox{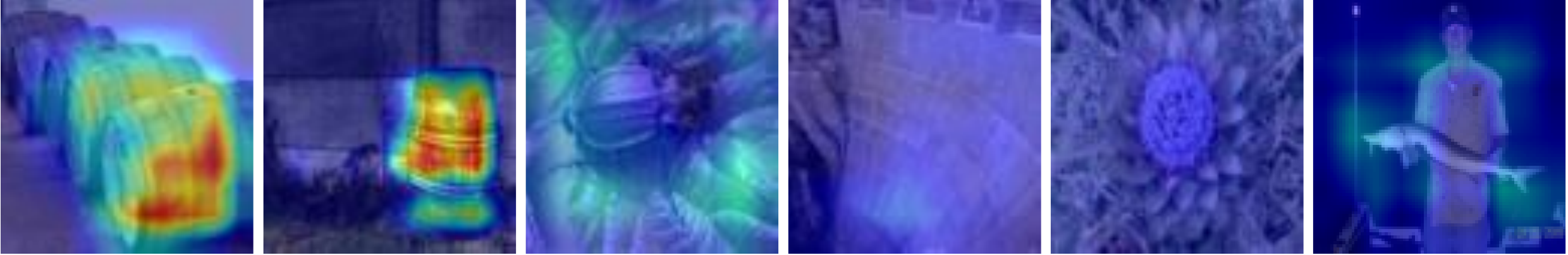}} & & \multicolumn{5}{c}{\cellbox{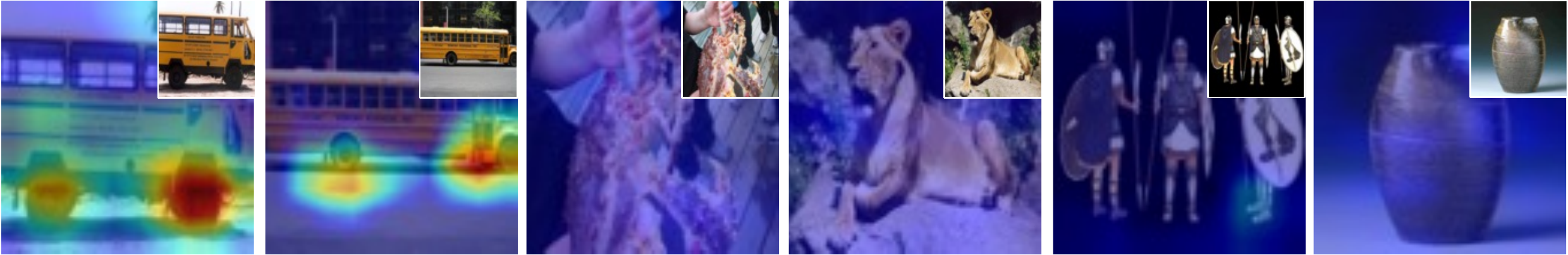}} \\
\multicolumn{1}{c}{} & \hspace{5pt}\scriptsize{query} & \scriptsize{ground truth}   & \multicolumn{3}{c}{\scriptsize{confounding support classes}} & & \hspace{6pt}\scriptsize{query}  & \scriptsize{ground truth}  & \multicolumn{3}{c}{\scriptsize{confounding support classes}} \\
\multicolumn{6}{c}{\small{(a) MCL as a plugin for feature centralization}} & & \multicolumn{5}{c}{\small{(b) MCL as an end-to-end classification method}}
\end{tabular}
\caption{Grad-CAM visualizations of query and support images in 5-way 1-shot tasks. The left pane (a) illustrates MCL as a plugin on the ProtoNet. The right pane (b) illustrates MCL as an end-to-end classification method. Images (from \textit{mini}ImageNet) at the second column of each pane are from the ground truth class while the four on the far right are the confounding support classes.}
    \label{tab:MEL_Quality}
\end{table*}

\textbf{Influence of Katz attenuation factor $\alpha$.} When $\alpha$ is small, the contribution given by paths longer than one rapidly declines, and the centrality will be determined by short paths (mostly in-degrees). When $\alpha$ is large, long paths are devalued smoothly, and the centrality would be more influenced by endogenous topology. When $\alpha$ approaching 0, the classification will be only determined by the unidirectional query-to-support random walk where the contributions from paths longer than one just vanished. We show comparisons in Table \ref{tab:uni} to demonstrate that performances are improved by introducing mutual affiliations in a bidirectional manner. 

In the experiments, we simply use $\alpha=0.5$ for MCL-Katz methods, and the performances could be further improved by selecting different $\alpha$ as shown in supplementary.

\subsection{Computational Speed}

Intuitively, bidirectional methods would be slower than unidirectional ones since their back-and-forth computations. However, our MCL would definitely reach to the periodic equilibrium state as shown in Eqn.(\ref{eq:stationary}). Thus, we could directly solve it by the Katz approximation. 

At first glance, the most expensive step in Eqn.(\ref{eq:katz_c}) is inverting $\bm{I} - \alpha\mathbf{P}$ that costs $O(N^3r^3)$ where $N$ is the number of supporting classes and $r$ is the number of dense features for each image input. Fortunately, $\bm{I} - \alpha\mathbf{P}$ is a very special matrix that equals a diagonal matrix minus an anti-diagonal matrix. In this case, we can reformulate it by block-wise matrix inversion:
\begin{equation}\label{eq:block}
\resizebox{0.85\linewidth}{!}{%
$\begin{aligned}[b]
\renewcommand{\arraystretch}{1.2}
    (\bm{I} - \alpha\mathbf{P})^{-1} &= \begin{pmatrix}
    \bm{I} & -\alpha \mathbf{P}_\mathbf{Sq} \\
    -\alpha \mathbf{P}_\mathbf{Sq} & \bm{I}
    \end{pmatrix}^{-1} \\
    &= \begin{pmatrix}
    \bm{I} + \alpha^2 \mathbf{P}_\mathbf{Sq}\mathbf{\Delta}^{-1}\mathbf{P}_\mathbf{qS} & \alpha \mathbf{P}_\mathbf{Sq}\mathbf{\Delta}^{-1} \\
    \alpha \mathbf{\Delta}^{-1} \mathbf{P}_\mathbf{qS} & \mathbf{\Delta}^{-1}\\
    \end{pmatrix}
\end{aligned}
$}
\end{equation}
where $\mathbf{\Delta}$ is defined by $\bm{I}-\alpha^2 \mathbf{P_{Sq}}\mathbf{P_{qS}}$. The most expensive step then becomes a inversion of $\mathbf{\Delta} \in \mathbb{R}^{r\times r}$ and the time complexity is reduced to $O(r^3)$.

As shown in Table \ref{tab:scale}, we compare the inference speed between MCL and other unidirectional dense feature based methods with different image resolutions on 5-way 1-shot FSL tasks with ResNet-12. Each class contains 15 query images in an episode. It can be observed that our bidirectional MCL did not introduce many computational overheads compared to unidirectional methods, and Katz approximation does accelerate computing the stationary distribution.

\subsection{Qualitative Feature Centralization} 

To better understand our centrality based methods, we visualize the Grad-CAM \cite{selvaraju2017grad} of the last convolutional layer in ResNet-12 for both plugin and end-to-end methods. We show in Table \ref{tab:MEL_Quality}(a) that our MCL helps ProtoNet concentrate on the most relevant regions of interest. Table \ref{tab:MEL_Quality}(b) demonstrates that the most centralized support features in MCL are not only from the ground truth but also mutually affiliated with task-relevant objects in the query images. We consider it qualitatively validates our underlying idea that support features would be frequently visited in the long times of bidirectional random walks if they shared the same local characteristics with the query image.

\begin{table}[t]{
\centering
\small
\aboverulesep=0ex
\belowrulesep=0ex
\renewcommand{\arraystretch}{1}
\setlength{\tabcolsep}{5pt}
\begin{tabular}{l|cccc}
\multirow{2}{*}{Model} & \multicolumn{4}{c}{\textit{VanillaFCN} resolution $r$} \\
& 5$\times$5 & 8$\times$8 & 10$\times$10 & 12$\times$12 \\ \toprule
DN4 \cite{li2019revisiting} & 11.16 & 13.41 & 16.82 & 23.45 \\
DeepEMD \cite{zhang2020deepemd} & 262.0 & 814.6 & 2781.9 & 8035.4 \\
FRN \cite{wertheimer2021few} \scriptsize{(Ridge regression)} & 14.55 & 20.42 & 25.81 & 30.65 \\
FRN \cite{wertheimer2021few} \scriptsize{(Woodbury identity)} & 36.74 & 42.33 & 47.31 & 55.84 \\ \midrule
MCL \scriptsize{(\texttt{torch.pinverse})} & 55.91 & 119.0 & 292.2 & 464.8 \\
MCL \scriptsize{(QR decomposition)} & 600.0 & 658.6 & 719.4 & 802.6 \\
MCL \scriptsize{(Katz approximation)} & 12.59 & 14.59 & 18.92 & 29.10
\end{tabular}
\caption{Time per episode (ms) for different feature map resolutions $r$ on 5-way 1-shot tasks where the feature extraction time is excluded for all comparing methods. \label{tab:scale}}}
\end{table}

\section{Conclusions}

We present a novel dense feature based framework: Mutual Centralized Learning (MCL) to highlight the mutual affiliations of bipartite dense features in FSL. We introduce a novel dense features' accessibility criterion and demonstrate classic transductive methods like label propagation could be easily adapted to the inductive setting if we treat dense features from a single image as the set of unlabeled data. We propose bidirectional random walk to learn mutual affiliations in FSL and prove its features' accessibility in a long time-homogeneous Markov process is equivalent to the single-mode eigenvector centrality of an affiliation network. We show that such centrality could not only serve as a noval end-to-end classification criterion but also as a plugin in existing methods. Experimental results demonstrate MCL achieves the state-of-the-art on various datasets.

{\small
\bibliographystyle{ieee_fullname}
\bibliography{egbib}

\begin{thebibliography}{10}\itemsep=-1pt

\bibitem{afrasiyabi2020associative}
Arman Afrasiyabi, Jean-Fran{\c{c}}ois Lalonde, and Christian Gagn{'e}.
\newblock Associative alignment for few-shot image classification.
\newblock In {\em European Conference on Computer Vision}, pages 18--35.
  Springer, 2020.

\bibitem{bertinetto2018meta}
Luca Bertinetto, Joao~F Henriques, Philip Torr, and Andrea Vedaldi.
\newblock Meta-learning with differentiable closed-form solvers.
\newblock In {\em International Conference on Learning Representations}, 2018.

\bibitem{bonacich1972factoring}
Phillip Bonacich.
\newblock Factoring and weighting approaches to status scores and clique
  identification.
\newblock {\em Journal of mathematical sociology}, 2(1):113--120, 1972.

\bibitem{bonacich1991simultaneous}
Phillip Bonacich.
\newblock Simultaneous group and individual centralities.
\newblock {\em Social networks}, 13(2):155--168, 1991.

\bibitem{borgatti1997network}
Stephen~P Borgatti and Martin~G Everett.
\newblock Network analysis of 2-mode data.
\newblock {\em Social networks}, 19(3):243--269, 1997.

\bibitem{chen2019closer}
Wei-Yu Chen, Yen-Cheng Liu, Zsolt Kira, Yu-Chiang~Frank Wang, and Jia-Bin
  Huang.
\newblock A closer look at few-shot classification.
\newblock In {\em International Conference on Learning Representations}, 2019.

\bibitem{chen2020new}
Yinbo Chen, Xiaolong Wang, Zhuang Liu, Huijuan Xu, and Trevor Darrell.
\newblock A new meta-baseline for few-shot learning.
\newblock {\em arXiv preprint arXiv:2003.04390}, 2020.

\bibitem{dhillon2019baseline}
Guneet~Singh Dhillon, Pratik Chaudhari, Avinash Ravichandran, and Stefano
  Soatto.
\newblock A baseline for few-shot image classification.
\newblock In {\em International Conference on Learning Representations}, 2019.

\bibitem{ding2020graph}
Kaize Ding, Jianling Wang, Jundong Li, Kai Shu, Chenghao Liu, and Huan Liu.
\newblock Graph prototypical networks for few-shot learning on attributed
  networks.
\newblock In {\em Proceedings of the 29th ACM International Conference on
  Information \& Knowledge Management}, pages 295--304, 2020.

\bibitem{doersch2020crosstransformers}
Carl Doersch, Ankush Gupta, and Andrew Zisserman.
\newblock Crosstransformers: spatially-aware few-shot transfer.
\newblock {\em arXiv preprint arXiv:2007.11498}, 2020.

\bibitem{garcia2017few}
Victor Garcia and Joan Bruna.
\newblock Few-shot learning with graph neural networks.
\newblock {\em arXiv preprint arXiv:1711.04043}, 2017.

\bibitem{hao2019collect}
Fusheng Hao, Fengxiang He, Jun Cheng, Lei Wang, Jianzhong Cao, and Dacheng Tao.
\newblock Collect and select: Semantic alignment metric learning for few-shot
  learning.
\newblock In {\em Proceedings of the IEEE/CVF International Conference on
  Computer Vision}, pages 8460--8469, 2019.

\bibitem{katz1953new}
Leo Katz.
\newblock A new status index derived from sociometric analysis.
\newblock {\em Psychometrika}, 18(1):39--43, 1953.

\bibitem{lee2019meta}
Kwonjoon Lee, Subhransu Maji, Avinash Ravichandran, and Stefano Soatto.
\newblock Meta-learning with differentiable convex optimization.
\newblock In {\em Proceedings of the IEEE/CVF Conference on Computer Vision and
  Pattern Recognition}, pages 10657--10665, 2019.

\bibitem{li2021asymmetric}
Wenbin Li, Lei Wang, Jing Huo, Yinghuan Shi, Yang Gao, and Jiebo Luo.
\newblock Asymmetric distribution measure for few-shot learning.
\newblock In {\em Proceedings of the Twenty-Ninth International Conference on
  International Joint Conferences on Artificial Intelligence}, pages
  2957--2963, 2021.

\bibitem{li2019revisiting}
Wenbin Li, Lei Wang, Jinglin Xu, Jing Huo, Yang Gao, and Jiebo Luo.
\newblock Revisiting local descriptor based image-to-class measure for few-shot
  learning.
\newblock In {\em Proceedings of the IEEE/CVF Conference on Computer Vision and
  Pattern Recognition}, pages 7260--7268, 2019.

\bibitem{lifchitz2021local}
Yann Lifchitz, Yannis Avrithis, and Sylvaine Picard.
\newblock Local propagation for few-shot learning.
\newblock In {\em 2020 25th International Conference on Pattern Recognition
  (ICPR)}, pages 10457--10464. IEEE, 2021.

\bibitem{lifchitz2019dense}
Yann Lifchitz, Yannis Avrithis, Sylvaine Picard, and Andrei Bursuc.
\newblock Dense classification and implanting for few-shot learning.
\newblock In {\em Proceedings of the IEEE/CVF Conference on Computer Vision and
  Pattern Recognition}, pages 9258--9267, 2019.

\bibitem{liu2020negative}
Bin Liu, Yue Cao, Yutong Lin, Qi Li, Zheng Zhang, Mingsheng Long, and Han Hu.
\newblock Negative margin matters: Understanding margin in few-shot
  classification.
\newblock In {\em European Conference on Computer Vision}, pages 438--455.
  Springer, 2020.

\bibitem{liu2019learning}
Yanbin {Liu}, Juho {Lee}, Minseop {Park}, Saehoon {Kim}, Eunho {Yang}, Sung~Ju
  {Hwang}, and Yi {Yang}.
\newblock Learning to propagate labels: Transductive propagation network for
  few-shot learning.
\newblock In {\em International Conference on Learning Representations}, 2019.

\bibitem{perron1907theorie}
Oskar Perron.
\newblock Zur theorie der matrices.
\newblock {\em Mathematische Annalen}, 64(2):248--263, 1907.

\bibitem{ren2018meta}
Mengye Ren, Eleni Triantafillou, Sachin Ravi, Jake Snell, Kevin Swersky,
  Joshua~B Tenenbaum, Hugo Larochelle, and Richard~S Zemel.
\newblock Meta-learning for semi-supervised few-shot classification.
\newblock In {\em International Conference on Learning Representations}, 2018.

\bibitem{sachin2017optimization}
Ravi Sachin and Larochell Hugo.
\newblock Optimization as a model for few-shot learning.
\newblock {\em ICLR}, 2017.

\bibitem{selvaraju2017grad}
Ramprasaath~R Selvaraju, Michael Cogswell, Abhishek Das, Ramakrishna Vedantam,
  Devi Parikh, and Dhruv Batra.
\newblock Grad-cam: Visual explanations from deep networks via gradient-based
  localization.
\newblock In {\em Proceedings of the IEEE international conference on computer
  vision}, pages 618--626, 2017.

\bibitem{simon2020adaptive}
Christian Simon, Piotr Koniusz, Richard Nock, and Mehrtash Harandi.
\newblock Adaptive subspaces for few-shot learning.
\newblock In {\em CVPR}, pages 4136--4145, 2020.

\bibitem{snell2017prototypical}
Jake Snell, Kevin Swersky, and Richard Zemel.
\newblock Prototypical networks for few-shot learning.
\newblock In {\em NIPS}, pages 4077--4087, 2017.

\bibitem{sun2019meta}
Qianru Sun, Yaoyao Liu, Tat-Seng Chua, and Bernt Schiele.
\newblock Meta-transfer learning for few-shot learning.
\newblock In {\em Proceedings of the IEEE/CVF Conference on Computer Vision and
  Pattern Recognition}, pages 403--412, 2019.

\bibitem{sung2018learning}
Flood Sung, Yongxin Yang, Li Zhang, Tao Xiang, Philip~HS Torr, and Timothy~M
  Hospedales.
\newblock Learning to compare: Relation network for few-shot learning.
\newblock In {\em CVPR}, pages 1199--1208, 2018.

\bibitem{tseng2020cross}
Hung-Yu Tseng, Hsin-Ying Lee, Jia-Bin Huang, and Ming-Hsuan Yang.
\newblock Cross-domain few-shot classification via learned feature-wise
  transformation.
\newblock In {\em International Conference on Learning Representations}, 2020.

\bibitem{vinyals2016matching}
Oriol Vinyals, Charles Blundell, Timothy Lillicrap, Koray Kavukcuoglu, and Daan
  Wierstra.
\newblock Matching networks for one shot learning.
\newblock In {\em Proceedings of the 30th International Conference on Neural
  Information Processing Systems}, pages 3637--3645, 2016.

\bibitem{welinder2010caltech}
Peter {Welinder}, Steve {Branson}, Takeshi {Mita}, Catherine {Wah}, Florian
  {Schroff}, Serge {Belongie}, and Pietro {Perona}.
\newblock Caltech-ucsd birds 200.
\newblock 2010.

\bibitem{wertheimer2019few}
Davis Wertheimer and Bharath Hariharan.
\newblock Few-shot learning with localization in realistic settings.
\newblock In {\em Proceedings of the IEEE/CVF Conference on Computer Vision and
  Pattern Recognition}, pages 6558--6567, 2019.

\bibitem{wertheimer2021few}
Davis Wertheimer, Luming Tang, and Bharath Hariharan.
\newblock Few-shot classification with feature map reconstruction networks.
\newblock In {\em Proceedings of the IEEE/CVF Conference on Computer Vision and
  Pattern Recognition}, pages 8012--8021, 2021.

\bibitem{ye2020few}
Han-Jia Ye, Hexiang Hu, De-Chuan Zhan, and Fei Sha.
\newblock Few-shot learning via embedding adaptation with set-to-set functions.
\newblock In {\em CVPR}, pages 8808--8817, 2020.

\bibitem{zhang2020deepemd}
Chi Zhang, Yujun Cai, Guosheng Lin, and Chunhua Shen.
\newblock Deepemd: Differentiable earth mover's distance for few-shot learning.
\newblock {\em arXiv e-prints}, pages arXiv--2003, 2020.

\bibitem{zhou2003learning}
Dengyong {Zhou}, Olivier {Bousquet}, Thomas~N. {Lal}, Jason {Weston}, and
  Bernhard {Schölkopf}.
\newblock Learning with local and global consistency.
\newblock In {\em Advances in Neural Information Processing Systems 16},
  volume~16, pages 321--328, 2003.

\end{thebibliography}
}

\clearpage

\renewcommand\thesection{\Alph{section}}

\appendix

\section{Proof of Eqn.(5)\label{sec:proof1}}

\paragraph{Proof of periodic:}

Consider the eigenvalue $\lambda$ of $\mathbf{P}$ by the determinant equation
\begin{equation*}\label{eq:periodic}
\begin{aligned}[b]
    \mathrm{det}(\lambda \bm{I} - \mathbf{P}) &= \mathrm{det} \begin{pmatrix}
    \lambda\bm{I} & -\mathbf{P}_{\mathbf{S}\mathbf{q}} \\
    -\mathbf{P}_{\mathbf{q}\mathbf{S}} & \lambda \bm{I}
\end{pmatrix} \\
&= \mathrm{det}(\lambda^2 \bm{I} - \mathbf{P}_\mathbf{Sq}\mathbf{P}_\mathbf{qS}),
\end{aligned}\tag{\ref{sec:proof1}.1}
\end{equation*}
it can be found that the eigenvalues of $\mathbf{P}$ are the square roots of eigenvalues of $\mathbf{P}_\mathbf{Sq}\mathbf{P}_\mathbf{qS}$.

Since both $\mathbf{P}_\mathbf{Sq}$ and $\mathbf{P}_\mathbf{qS}$ are column-normalized matrices, their product is still column-stochastic that can be proved by:
\begin{equation*}\label{eq:stoch}
 \bm{e}_{Nr}^\mathrm{T}\mathbf{P}_\mathbf{Sq}\mathbf{P}_\mathbf{qS}=\bm{e}_{r}^\mathrm{T}\mathbf{P}_\mathbf{qS}=\bm{e}_{Nr}^\mathrm{T}\tag{\ref{sec:proof1}.2}
\end{equation*}
where $\bm{e}_{|\cdot|}$ is a vector of ones with different length indicated by its subscript. $Nr$ and $r$ are the cardinalities of $\mathbf{S}$ and $\mathbf{q}$, respectively.

We know (by the definition of stochastic matrix) that $\lambda=1$ is the largest eigenvalue of $\mathbf{P}_\mathbf{Sq}\mathbf{P}_\mathbf{qS}$, and its uniqueness is guaranteed since there is no zero entry in both $\mathbf{P}_\mathbf{Sq}$ and $\mathbf{P}_\mathbf{qS}$. According to Eqn.(\ref{eq:periodic}), we get another eigenvalue $\lambda=-1$ for stochastic matrix $\mathbf{P}$. From the Perron–Frobenius theorem that the period of $\mathbf{P}$ equals to the number of eigenvalue whose absolute value is equal to the spectral radius of $\mathbf{P}$, we prove its stationary distribution is of period 2.

\noindent\paragraph{Proof for even periods:}

We give the limit of matrix $\mathbf{P}^{2t}$ for the extremely large number of $t$ as follows:
\begin{equation*}\label{eq:even_proof}
\begin{aligned}
    \lim_{t\rightarrow\infty} \mathbf{P}^{2t} &= \lim_{t\rightarrow\infty} \begin{pmatrix}
      \mathbf{P}_\mathbf{Sq}\mathbf{P}_\mathbf{qS} & \mathbf{0} \\
     \mathbf{0} & \mathbf{P}_\mathbf{qS}\mathbf{P}_\mathbf{Sq}
    \end{pmatrix}^t \\
    &= \begin{pmatrix}
     \displaystyle{\lim_{t\rightarrow\infty}} \left[\mathbf{P}_\mathbf{Sq}\mathbf{P}_\mathbf{qS}\right]^t & \mathbf{0} \\
     \mathbf{0} & \displaystyle{\lim_{t\rightarrow\infty}} \left[\mathbf{P}_\mathbf{qS}\mathbf{P}_\mathbf{Sq}\right]^t
    \end{pmatrix}
\end{aligned}\tag{\ref{sec:proof1}.3}
\end{equation*}

Since we have shown in Eqn.(\ref{eq:stoch}) that $\mathbf{P}_\mathbf{Sq}\mathbf{P}_\mathbf{qS}$ is also column-stochastic, we use $\boldsymbol{\pi}(\mathbf{S})$ to denote its stationary distribution vector by equation $\displaystyle{\lim_{t\rightarrow\infty}} \left[\mathbf{P}_\mathbf{Sq}\mathbf{P}_\mathbf{qS}\right]^t = \boldsymbol{\pi}(\mathbf{S})\bm{e}_{Nr}^\mathrm{T}$. 

By analogy, the infinity power of $\mathbf{P}_\mathbf{qS}\mathbf{P}_\mathbf{Sq}$ could also reach a similar stationary distribution $\boldsymbol{\pi}(\mathbf{q})$ with equation $\displaystyle{\lim_{t\rightarrow\infty}} \left[\mathbf{P}_\mathbf{qS}\mathbf{P}_\mathbf{Sq}\right]^t = \boldsymbol{\pi}(\mathbf{q})\bm{e}_{r}^\mathrm{T}$. 

Substituting the two stationary vectors into Eqn.(\ref{eq:even_proof}), we can prove the stationary distributions of $\mathbf{P}$ for the even periods in Eqn.(5).

\noindent\paragraph{Proof for odd periods:}

From the definition of matrix product, we first have
\begin{equation*}\label{eq:odd_def}
\begin{aligned}
    \lim_{t\rightarrow\infty} \mathbf{P}^{2t-1} &= \lim_{t\rightarrow\infty} \mathbf{P}^{2t+1} \\ 
    &= \mathbf{P}\lim_{t\rightarrow\infty} \mathbf{P}^{2t} \\
    &= \begin{pmatrix}
    \mathbf{0} & \mathbf{P}_\mathbf{Sq} \boldsymbol{\pi}(\mathbf{q})\bm{e}_{r}^\mathrm{T} \\ \mathbf{P}_\mathbf{qS} \boldsymbol{\pi}(\mathbf{S})\bm{e}_{Nr}^\mathrm{T} & \mathbf{0}
    \end{pmatrix}
\end{aligned}\tag{\ref{sec:proof1}.4}
\end{equation*}

Next, according to the definition of $\boldsymbol{\pi}(\mathbf{q})$ and $\boldsymbol{\pi}(\mathbf{S})$, we can get
\begin{equation*}\label{eq:B.5}
\begin{aligned}
    \boldsymbol{\pi}(\mathbf{q})\bm{e}_{r}^\mathrm{T} &= \lim_{t\rightarrow\infty}\left[\mathbf{P}_\mathbf{qS}\mathbf{P}_\mathbf{Sq}\right]^t \\
    &= \mathbf{P}_\mathbf{qS}\left(\lim_{t\rightarrow\infty}\left[\mathbf{P}_\mathbf{Sq}\mathbf{P}_\mathbf{qS}\right]^t\right)\mathbf{P}_\mathbf{Sq} \\
    &= \mathbf{P}_\mathbf{qS}\boldsymbol{\pi}(\mathbf{S})\bm{e}_{Nr}^\mathrm{T}\mathbf{P}_\mathbf{Sq}.
\end{aligned}
\tag{\ref{sec:proof1}.5}
\end{equation*}
If we right matrix product of $\bm{e}_{r}$ on both sides of Eqn.(\ref{eq:B.5}), we have
\begin{equation*}\label{eq:B.6}
    \boldsymbol{\pi}(\mathbf{q})\bm{e}_{r}^\mathrm{T}\bm{e}_{r} = \mathbf{P}_\mathbf{qS}\boldsymbol{\pi}(\mathbf{S})\bm{e}_{Nr}^\mathrm{T}\mathbf{P}_\mathbf{Sq}\bm{e}_{r} \tag{\ref{sec:proof1}.6}
\end{equation*}

Since $\bm{e}_{r}^\mathrm{T}\bm{e}_{r} = r$ and $\bm{e}_{Nr}^\mathrm{T}\mathbf{P}_\mathbf{Sq}\bm{e}_{r} = {\sum_i\sum_j\left[\mathbf{P}_\mathbf{Sq}\right]_{ij}}=r$, Eqn.(\ref{eq:B.6}) can be simplified by dividing the same scalar $r$ on both sides:
\begin{equation*}\label{eq:B.7}
    \boldsymbol{\pi}(\mathbf{q}) = \mathbf{P}_\mathbf{qS}\boldsymbol{\pi}(\mathbf{S}).\tag{\ref{sec:proof1}.7}
\end{equation*}

By analogy, a symmetric equation $\boldsymbol{\pi}(\mathbf{S}) = \mathbf{P}_\mathbf{Sq}\boldsymbol{\pi}(\mathbf{q})$ can also be easily proved in the same way as from Eqn.(\ref{eq:B.5}) to Eqn.(\ref{eq:B.7}). 

Substituting $\boldsymbol{\pi}(\mathbf{S}) = \mathbf{P}_\mathbf{Sq}\boldsymbol{\pi}(\mathbf{q})$ and $\boldsymbol{\pi}(\mathbf{q}) = \mathbf{P}_\mathbf{qS}\boldsymbol{\pi}(\mathbf{S})$ into Eqn.(\ref{eq:odd_def}), we can prove the stationary distributions of $\mathbf{P}$ for the odd periods in Eqn.(5).

\section{Proof of Lemma 1\label{sec:proof2}}

We have shown in Appendix \ref{sec:proof1} that there exists an eigenvalue $\lambda=1$ for the column-stochastic matrix $\mathbf{P}$ with equation $\mathbf{Px}= \mathbf{x}$. If we interpret the transition matrix as an adjacency matrix for the directed bipartite graph, the eigenvector centrality of that graph is $\mathbf{x}$.

We split the eigenvector $\mathbf{x}$ into $\mathbf{x}_\mathbf{S}$, $\mathbf{x}_\mathbf{q}$ for the bipartite vertex set $\mathbf{q}, \mathbf{S}$ respectively and the single-mode eigenvector centralities of the single vertex set can therefore be formulated by:
\begin{align*}\label{eq:single}
    \hat{\mathbf{x}}_{\mathbf{S}}&= \frac{\mathbf{x}_\mathbf{S}}{\sum_{s\in \mathbf{S}}x_s} & \hat{\mathbf{x}}_{\mathbf{q}}&=\frac{\mathbf{x}_\mathbf{q}}{\sum_{q\in \mathbf{q}}x_q}\tag{\ref{sec:proof2}.1}
\end{align*}

If we left matrix product $\mathbf{P}$ on both sides of $\mathbf{Px}=\mathbf{x}$, we will have $\mathbf{P}^2\mathbf{x}=\mathbf{P}(\mathbf{Px})=\mathbf{Px}=\mathbf{x}$. To write it in matrix notation, we have
\begin{equation}\label{eq:C.2}
    \renewcommand{\arraystretch}{1.5}
    \underbrace{\begin{pmatrix}
    \mathbf{P}_\mathbf{Sq}\mathbf{P}_\mathbf{qS} & \mathbf{0} \\
    \mathbf{0} & \mathbf{P}_\mathbf{qS}\mathbf{P}_\mathbf{Sq}
\end{pmatrix}}_{\mathbf{P}^2} \underbrace{\begin{pmatrix} \mathbf{x}_\mathbf{S} \\ \mathbf{x}_\mathbf{q} 
\end{pmatrix}}_\mathbf{x} = \underbrace{\begin{pmatrix} \mathbf{x}_\mathbf{S} \\ \mathbf{x}_\mathbf{q} 
\end{pmatrix}}_\mathbf{x}. \tag{\ref{sec:proof2}.2}
\end{equation}

Consider the first row of $\mathbf{P}^2$ matrix product with $\mathbf{x}$ in Eqn.(\ref{eq:C.2}), we have $\mathbf{P}_\mathbf{Sq}\mathbf{P}_\mathbf{qS} \mathbf{x}_\mathbf{S} = \mathbf{x}_\mathbf{S}$. Since $\boldsymbol{\pi}(\mathbf{S})$ is the eigenvector of $\mathbf{P}_\mathbf{Sq}\mathbf{P}_\mathbf{qS}$ of eigenvalue 1 with probability constraint $\bm{e}_{Nr}^\mathrm{T}\boldsymbol{\pi}(\mathbf{S}) = 1$, $\boldsymbol{\pi}(\mathbf{S})$ is exactly equivalent to the single-mode eigenvector centrality $\hat{\mathbf{x}}_{\mathbf{S}}$ in Eqn.(\ref{eq:single}). 

By analogy, if we consider the matrix product between the second row of $\mathbf{P}^2$ and $\mathbf{x}$ in Eqn.(\ref{eq:C.2}), we can prove $\boldsymbol{\pi}(\mathbf{q})$ equivalent to the conjugate single-mode eigenvector centrality $\hat{\mathbf{x}}_\mathbf{q}$ of the bipartite graph.

\section{Proof of Eqn.(6)\label{sec:proof3}}

\noindent\textbf{Eqn.(6) to prove:}
\begin{equation*}
\begin{aligned}
    \mathbf{Pr}(\tilde{y}=c) &= \lim_{t\rightarrow\infty} \frac{\displaystyle{\sum_{z\in\mathbf{z}}}\hspace{2pt}\mathbb{E} \left[\sum_{k=1}^t\mathds{1}[X_k\in \mathbf{s}^c] \bigg| X_0=z\right]}{\displaystyle{\sum_{z\in\mathbf{z}}}\hspace{2pt}\mathbb{E} \left[\sum_{k=1}^t\mathds{1}[X_k\in \mathbf{S}] \bigg| X_0=z\right]} \\
    &= \sum_{s\in \mathbf{s}^c}\left[\boldsymbol{\pi}(\mathbf{S})\right]_s
\end{aligned}
\end{equation*}

We first define $\displaystyle{\mathbf{Pr}(\tilde{y}=c)\triangleq \lim_{t\rightarrow\infty}\mathbf{Pr}(t)}$. Since we have proof that Markov process of transition matrix $\mathbf{P}$ is of 2 period in Appendix \ref{sec:proof1}, the proof of Eqn.(6) is thus equivalent to prove: 
\begin{equation}
    \lim_{t\rightarrow\infty}\mathbf{Pr}(2t) = \lim_{t\rightarrow\infty}\mathbf{Pr}(2t - 1) = \sum_{s\in \mathbf{s}^c}\left[\boldsymbol{\pi}(\mathbf{S})\right]_s\tag{\ref{sec:proof3}.1}
\end{equation}

\paragraph{Proof for even period:}

From the definition, we have
\begin{align*}\label{eq:D.2}
\begin{split}
    &\mathbf{Pr}(2t) = \frac{\displaystyle{\frac{1}{Nr+r}\sum_{z\in \mathbf{z}}\sum_{k=1}^{2t}}\sum_{s\in \mathbf{s}^c}\left[\mathbf{P}^k\right]_{sz}}{\displaystyle{\frac{1}{Nr+r}\left(rt + Nrt\right)}} \\
    &= \frac{1}{t}\sum_{k=1}^{t}\left[\frac{1}{Nr+r}\sum_{s\in \mathbf{s}^c}\left(\sum_{z\in \mathbf{S}}\left[\mathbf{P}^{2k}\right]_{sz} + \sum_{z\in \mathbf{q}}\left[\mathbf{P}^{2k-1}\right]_{sz}\right)\right]
\end{split}\tag{\ref{sec:proof3}.2}
\end{align*}
where $rt$ is the number of visits from particles in $\mathbf{q}$ to support features in $\mathbf{S}$ after $2t$ steps of Markov bidirectional random walk. $Nrt$ is the number of visits starting from particles in $\mathbf{S}$ to support features in $\mathbf{S}$. The second equality is derived from the diagonal/anti-diagonal property of $\mathbf{P}^{2k}$/$\mathbf{P}^{2k-1}$ respectively where the sub-matrices $\mathbf{0}$ are ignored in summation. 

Taking Eqn.(\ref{eq:D.2}) to the extreme, we have
\begin{equation*}\label{eq:D.3}
\begin{aligned}
    &\lim_{t\rightarrow\infty}\mathbf{Pr}(2t) \\ 
    &= \frac{1}{Nr+r} \sum_{s\in \mathbf{s}^c}\left(\sum_{z\in \mathbf{S}}\left[\lim_{t\rightarrow \infty}\mathbf{P}^{2t}\right]_{sz} + \sum_{z\in \mathbf{q}}\left[\lim_{t\rightarrow \infty}\mathbf{P}^{2t-1}\right]_{sz}\right) \\
    &= \sum_{s\in \mathbf{s}^c} \left[\boldsymbol{\pi}(\mathbf{S})\right]_s
\end{aligned}\tag{\ref{sec:proof3}.3}
\end{equation*}
where the first equality is derived from the \textit{absorbing} of periodic Markov chain and the second equality is from the substitution of Eqn.(5).

\paragraph{Proof for odd period}: From the definition, we have
\begin{align*}\label{eq:D.4}
\begin{split}
    &\mathbf{Pr}(2t-1) = \frac{\displaystyle{\frac{1}{Nr + r}\sum_{z\in \mathbf{z}}\sum_{k=1}^{2t-1}}\sum_{s\in \mathbf{s}^c}\left[\mathbf{P}^k\right]_{sz}}{{\displaystyle{\frac{1}{Nr + r}\left(rt + Nr(t - 1)\right)}}} \\
    &= \frac{1}{\omega}\sum_{s\in \mathbf{s}^c}\left[\sum_{z\in \mathbf{q}}\mathbf{P}_{sz} + \sum_{k=2}^{t}\left(\sum_{z\in \mathbf{S}}\left[\mathbf{P}^{2k-2}\right]_{sz} + \sum_{z\in \mathbf{q}}\left[\mathbf{P}^{2k-1}\right]_{sz}\right)\right]
\end{split}\tag{\ref{sec:proof3}.4}
\end{align*}
where $\omega$ equals $(Nr + r)(t - \frac{Nr}{Nr + r})$. Take Eqn.(\ref{eq:D.4}) to the extreme, we have
\begin{align*}
\begin{split}
    &\lim_{t\rightarrow\infty}\mathbf{Pr}(2t-1) \\
    &= \frac{1}{Nr+r} \sum_{s\in \mathbf{s}^c}\left(\sum_{z\in \mathbf{S}}\left[\lim_{t\rightarrow \infty}\mathbf{P}^{2t-2}\right]_{sz} + \sum_{z\in \mathbf{q}}\left[\lim_{t\rightarrow \infty}\mathbf{P}^{2t-1}\right]_{sz}\right) \\
    &= \sum_{s\in \mathbf{s}^c} \left[\boldsymbol{\pi}(\mathbf{S})\right]_s
\end{split}\tag{\ref{sec:proof3}.5}
\end{align*}
where $\displaystyle{\lim_{t\rightarrow\infty}\frac{1}{t - \frac{Nr}{Nr +r}}\sum_{z\in\mathbf{q}}\mathbf{P}_{sz}} = 0$ is ignored when $t$ approaches the infinity. 

\renewcommand{\lstlistingname}{Code}
\begin{figure*}[h]
\begin{minipage}{1.0\textwidth}
\begin{lstlisting}[language=Python, caption={Pytorch pseudo-code for 1-shot MCL (Katz approximation) in a single episode.},label={code}]
# support of tensor shape [N, d, r]:
#     N-way FSL, each class owns r number of d-dimensional dense features
# query of tensor shape [q, d, r]: 
#     q query examples, each of them owns r dense features.
#
# gamma: scaled similarity parameter
# beta: scaled similarity parameter
# alpha: Katz attenuation factor
# alpha_2: the square of alpha
# 
# @: the matrix multiplication operator in Pytorch
def inner_cosine(query, support):
    N, d, r = support.shape
    q = len(query)
    query = query / query.norm(2, dim=-1, keepdim=True)
    support = support / support.norm(2, dim=-1, keepdim=True)
    
    support = support.unsqueeze(0).expand(q, -1, -1, -1)
    query = query.unsqueeze(1).expand(-1, N, -1, -1)
    S = query.transpose(-2, -1)@support
    S = S.permute(0, 2, 1, 3).contiguous().view(q, r, N * r)
    return S

def MCL_Katz_approx(query, support):
    N, d, r = support.shape
    q = len(query)
    S = inner_cosine(query, support) # [q, r, Nr]
    St = S.transpose(-2, -1) # [q, Nr, r]
    
    # column-wise softmax probability
    P_sq = torch.softmax(gamma * St, dim=-2)
    P_qs = torch.softmax(beta * S, dim=-2)
    # From the derivations in Eqn.(F.2)
    inv = torch.inverse(
        torch.eye(r)[None].repeat(q, 1, 1) - alpha_2 * P_qs@P_sq
    ) # [q, r, r]
    katz = (alpha_2 * P_sq@inv@P_qs).sum(-1) + (alpha * P_sq@inv).sum(-1)
    katz = katz / katz.sum(-1, keepdim=True)
    predicts = katz.view(q, N, r).sum(-1)
    return predicts
\end{lstlisting}
\end{minipage}
\end{figure*}

\section{Codes}

We use block-wise inversion in Eqn.(14) that is more computational-efficient than directly inverting Eqn.(10) when the number of support classes $N$ is large. The corresponding Pytorch pseudo code can be found below where the whole calculation is performed in parallel via batched matrix multiplication and inversion. Our unified FSL training and testing framework is publicly available at \url{https://github.com/LouieYang/MCL}. 

\section{Implementation details}

\textbf{Preprocessing:}
During training on CUB, \textit{mini}ImageNet and \textit{tiered}ImageNet, images are randomly cropped to $92\times 92$ and then resized into $84\times 84$. For meta-iNat and \textit{tiered}-meta-iNat, images are randomly padded then cropped into $84\times 84$. Unlike previous methods , we only random horizontal flip the image during training.

During inference, images are center cropped to $92\times 92$ and then resized into $84\times 84$ for CUB, \textit{mini}ImageNet and \textit{tiered}ImageNet. For meta-iNat and \textit{tiered}-meta-iNat, images are already $84\times 84$ and are fed into the models directly.

\textbf{Network backbones:} We use two backbones in our experiments: Conv-4 and ResNet-12. Conv-4 contains four convolutional blocks, each of which consists of a $3\times 3$ convolution, a BatchNorm, a LeakyReLU(0.2) and an additional $2\times 2$ max-pooling. ResNet-12 consist four residual blocks, each with three convolutional layers, with LeakyReLU(0.1) and $2\times 2$ max-pooling on the main stem. Given the image of input size $84\times 84$, Conv-4 outputs a feature map of size $5\times 5\times 64$ while ResNet-12 outputs that of size $5\times 5 \times 64$.

\textbf{Re-implemented baselines:} We re-implement ProtoNet \cite{snell2017prototypical} and RelationNet \cite{sung2018learning} in our unified framework as two global feature based baselines for our centrality plugin. We borrow most of codes from their official implementations and introduce slight modifications to improve their performances inspired their subsequent work \cite{wertheimer2021few, ye2020few}. For ProtNet, we introduce a fixed temperature scaling with $1/64$ before in softmax function. For RelationNet, we change the original MSELoss to CrossEntropyLoss.

\begin{table}[t]
\small
\centering
\setlength{\tabcolsep}{4pt}
\begin{tabular}{lccc}
\toprule
\textbf{Model} & \textbf{Backbone} & \textbf{1-shot} & \textbf{5-shot} \\ \midrule
Baseline \cite{chen2019closer} & ResNet-10 & - & 65.57 \scriptsize{$\pm$ 0.70} \\
Baseline++ \cite{chen2019closer} & ResNet-18 & - & 62.04 \scriptsize{$\pm$ 0.76} \\
MetaOptNet \cite{lee2019meta}& ResNet-12 & 44.79 \scriptsize{$\pm$ 0.75} & 64.98 \scriptsize{$\pm$ 0.68} \\
MatchingNet+FT \cite{tseng2020cross} & ResNet-10 & 36.61 \scriptsize{$\pm$ 0.53} & 55.23 \scriptsize{$\pm$ 0.83} \\
RelationNet+FT \cite{tseng2020cross} & ResNet-10 & 44.07 \scriptsize{$\pm$ 0.77} & 59.46 \scriptsize{$\pm$ 0.71} \\
GNN+FT \cite{tseng2020cross} & ResNet-10 & 47.47 \scriptsize{$\pm$ 0.75} & 66.98 \scriptsize{$\pm$ 0.68} \\
Neg-margin \cite{liu2020negative} & ResNet-18 & - & 69.30 \scriptsize{$\pm$ 0.73} \\
Centroid \etal \cite{afrasiyabi2020associative} & ResNet-18 & 46.85 \scriptsize{$\pm$ 0.75} & 70.37 \scriptsize{$\pm$ 1.02} \\
FRN \cite{wertheimer2021few}& ResNet-12 & 51.60 \scriptsize{$\pm$ 0.21} & 72.97 \scriptsize{$\pm$ 0.18} \\
\midrule
ProtoNet$^\dagger$ \cite{snell2017prototypical}& ResNet-12 & 40.05 \scriptsize{$\pm$ 0.18} & 55.29  \scriptsize{$\pm$ }0.19 \\ 
ProtoNet+MCL& ResNet-12 & 42.02 \scriptsize{$\pm$ 0.19} & 64.76 \scriptsize{$\pm$ 0.20} \\ 
MCL & ResNet-12 & \textbf{55.48} \scriptsize{$\pm$ \textbf{0.22}} & 75.93 \scriptsize{$\pm$ 0.18} \\
MCL-Katz & ResNet-12 & 53.22 \scriptsize{$\pm$ 0.22} & \textbf{77.39} \scriptsize{$\pm$ \textbf{0.18}} \\ \bottomrule
\end{tabular}
\caption{Few-shot classifications (\%) in the cross-domain setting: \textit{mini}ImageNet$\rightarrow$CUB. $\dagger$: our re-implemented results in our unified framework that share the same dataloader and training strategies.\label{tab:cross}}
\end{table}

\textbf{Pre-training:} We use the pre-training + meta-training procedure for ResNet-12 backbones on \textit{mini}ImageNet and \textit{tiered}ImageNet like most of the methods in the literature \cite{zhang2020deepemd, li2019revisiting, wertheimer2021few}. We follow the same pre-training technique from the dense features based FRN \cite{wertheimer2021few} to learn spatially distinctive dense features, \eg, run 350 epochs of batch size 128 on \textit{mini}ImageNet, using SGD with initial learning rate 0.1 and decaying by a factor of 10 at epochs 200 and 300.

\textbf{Meta-training on \textit{mini}-/\textit{tiered}ImageNet:} We follow the earliest dense feature based DN4 \cite{li2019revisiting} that randomly sample 20,000/200 episodes in an epoch for Conv-4/ResNet-12 backbone, respectively. Since we didn not use pre-training for Conv-4, the number of episodes per epoch of Conv-4 is far larger than ResNet-12 for convergences. In each episode, besides $K $ support images in each class, 15 query images will also be selected from each class. 

For Conv-4, we adopt Adam optimizer with initial learning rate of 1e-3 to train for 30 epochs (on both datasets) and reduce it by a factor of 10 every 10 epochs. 

For ResNet-12, we adopt SGD with initial learning rate of 5e-4 (40 epochs on \textit{mini}ImageNet and 60 epochs on \textit{tiered}ImageNet) and cut it by half every 10 epochs.

Unlike the latest dense feature based FRN that adopts larger way during meta-training (\eg, 25-way for training 1-shot models and 20-way for 5-shot models), we did not use that setting in widely used \textit{mini-}/\textit{tirered}ImageNet as we thought it would be unfair in comparing with other methods. 

\textbf{Meta-training on fine-grained datasets:} We follow the latest fine-grained few-shot classification setting from FRN \cite{wertheimer2021few} as most of the comparing performances in Table 2 are from them. Although practitioners agree on the train/validation/test split ratio (\ie, 100/50/50) on CUB dataset, there is no official class split. In our experiments, we use the same train/val/test class split as in \cite{wertheimer2021few} for a fair comparison.  

For CUB, we train all our Conv-4 models for 1200 epochs using SGD with Nesterov momentum of 0.9 and an initial learning rate of 0.1. The learning rate decreases by a factor of 10 at epoch 400 and 800. ResNet-12 backbone trains for 600 epochs and scale down the learning rate by 10 at epoch 300, 400 and 500. Conv-4 backbone is trained with standard 20-way for 5-shot models and is trained with 30-way for 1-shot models like \cite{wertheimer2021few}, while ResNet-12 backbone is trained with 10-way for the both shots models.

For meta-iNat and \textit{tiered}-meta-iNat, we train our Conv-4 and ResNet-12 models for 100 epochs using Adam with initial learning rate 1e-3. We set 0.5 learning rate decay every 20 epochs. Both Conv-4 and ResNet-12 are trained with 10-way for both 1-shot and 5-shot models.

\begin{table}[t]
\centering
\small
\setlength{\tabcolsep}{5pt}
\begin{tabular}{lccc}
\toprule
\textbf{Methods} & \textbf{$f_\theta(\cdot)$} & \textbf{1-shot} & \textbf{5-shot} \\ \midrule
Baseline \cite{chen2019closer} & \multirow{4}{*}{\textit{Global feature}} & 53.99 \scriptsize{$\pm$ 0.20} & 78.78 \scriptsize{$\pm$ 0.15} \\
Baseline++ \cite{chen2019closer} &  & 55.03 \scriptsize{$\pm$ 0.20} & 78.79 \scriptsize{$\pm$ 0.15}  \\
ProtoNet \cite{snell2017prototypical} &  & 53.99 \scriptsize{$\pm$ 0.20} & 79.70 \scriptsize{$\pm$ 0.15} \\
ProtoNet+MCL &  & 58.62 \scriptsize{$\pm$ 0.20} & 80.99 \scriptsize{$\pm$ 0.14}  \\ \hline
DN4 \cite{li2019revisiting} & \multirow{4}{*}{\textit{VanillaFCN}} & 58.95 \scriptsize{$\pm$ 0.19}  & 78.01 \scriptsize{$\pm$ 0.15} \\
FRN \cite{wertheimer2021few} &  & 59.71 \scriptsize{$\pm$ 0.19} & 74.05 \scriptsize{$\pm$ 0.15} \\
MCL &  & 60.94 \scriptsize{$\pm$ 0.20} & 80.20 \scriptsize{$\pm$ 0.14} \\
MCL-Katz &  & \textbf{61.55} \scriptsize{$\pm$ \textbf{0.20}} & \textbf{81.09} \scriptsize{$\pm$ \textbf{0.14}} \\
\bottomrule
\end{tabular}
\caption{Few-shot classifications (\%) without episodic meta-training. All of the comparing methods are evaluated in our unified framework with the same pre-trained ResNet-12 backbone. \label{tab:wometa}}
\end{table}

\section{Cross-Domain Few-shot Classification}

We also evaluate on the challenging cross-domain setting proposed by \cite{chen2019closer}, where models trained on \textit{mini}ImageNet base classes are directly evaluated on test classes from CUB. We use the same test class split as in \cite{wertheimer2021few} for fair comparisons, which is much harder than the test class split in \cite{chen2019closer}. 

As shown in Table S\ref{tab:cross}, our MCL outperforms previous state-of-the-art methods by large margins of \textbf{3.9}\% on the 1-shot task and \textbf{5.4}\% on the 5-shot task, respectively.

\section{Evaluation without Meta-training\label{sec:wometa}}

Given that an increasing number of methods simply use standard supervised learning to pre-train the feature extractor and then use their methods directly for evaluation without meta-training \cite{chen2019closer, dhillon2019baseline}, we also evaluate our methods under this setting with the same pre-trained feature extractor we used in the Table 1. As shown in Table S\ref{tab:wometa}, global feature based methods are likely to misclassify images under the extremely 1-shot scenario, where the significant intra-class variations would inevitably drive the image-level embedding from  the  same category far apart. In contrast, dense feature based methods provide more information across categories that shows promising performances in that scenario. 

Our end-to-end Katz centrality based MCL outperforms previous methods by a margin of \textbf{1.8}\% and \textbf{1.3}\% on 1-shot and 5-shot tasks, respectively. It is interested to note that our MCL plugin help centralize the task-relevant local features in ProtoNet \cite{snell2017prototypical} by a large margin of \textbf{4.6}\% in 1-shot task and \textbf{1.2}\% in 5-shot task without bell and whistles.

\begin{table}[t]
\centering
\small
\begin{tabular}{lcc}
\toprule
 & \textbf{5-way 1-shot} & \textbf{5-way 5-shot} \\ \midrule
$\alpha\approx 0$ (unidirectional) & 66.60 \scriptsize{$\pm$ 0.20} & 81.76 \scriptsize{$\pm$ 0.13} \\
$\alpha=0.1$ & 67.13 \scriptsize{$\pm$ 0.20} & 83.20 \scriptsize{$\pm$ 0.13} \\
$\alpha=0.3$ & 67.28 \scriptsize{$\pm$ 0.20} & 83.89 \scriptsize{$\pm$ 0.13} \\
$\alpha=0.5$ (MCL-Katz) & \textbf{67.51} \scriptsize{$\pm$ \textbf{0.20}} & 83.99 \scriptsize{$\pm$ 0.13} \\
$\alpha=0.7$ & 67.27 \scriptsize{$\pm$ 0.20} & \textbf{84.05} \scriptsize{$\pm$ \textbf{0.13}} \\
$\alpha=0.9$ & 67.41 \scriptsize{$\pm$ 0.20} & 83.96 \scriptsize{$\pm$ \textbf{0.13}} \\
$\alpha=0.999$ (MCL) & 67.36 \scriptsize{$\pm$ 0.20} & 83.63 \scriptsize{$\pm$ 0.13} \\ \bottomrule
\end{tabular}
\vspace{-3pt}
\caption{Ablation studies on the Katz attenuation factor $\alpha$ on \textit{mini}ImageNet with \textit{VanillaFCN} ResNet-12. \label{tab:ab_alpha}}
\end{table}

\begin{table}[t]
\small
\centering
\begin{tabular}{ccccc}
\toprule
 & $\beta=5.0$ & $\beta=10.0$ & $\beta=20.0$ & $\beta=50.0$ \\ \midrule
$\gamma=5.0$  & 66.12 & 66.15 & 65.87 & 65.39 \\
$\gamma=10.0$ & 67.21 & 67.20 & 66.78 & 66.23 \\
$\gamma=20.0$ & 67.53 & \textbf{67.36} & 66.70 & 66.00 \\
$\gamma=50.0$ & 67.05 & 66.77 & 65.84 & 65.08 \\
\bottomrule
\end{tabular}
\caption{Ablation studies of MCL ($\alpha=0.999$) on the scaling parameters $\gamma$ and $\beta$ in Eqn.(1) and Eqn.(2), respectively. Experiments are conducted on 5-way 1-shot \textit{mini}ImageNet with \textit{VanillaFCN}.\label{tab:ab_gamma}}
\end{table}

\section{Ablation on parameters $\alpha$, $\gamma$ and $\beta$}

\textbf{Ablation on Katz attenuation factor $\alpha$.} Table S\ref{tab:ab_alpha} shows results with different $\alpha$ in bidirectional random walks. As discussed in Sec.5, the centrality with large $\alpha$ is more influenced by the endogenous topology while that with small $\alpha$ is more influenced by the in-degree paths. It could be observed in Table S\ref{tab:ab_alpha} that, the optimal $\alpha$ differs across various FSL tasks (that with different shot on different datasets). 

In the experiments, we use $\alpha$\texttt{=}$0.999$ to approximate single-mode eigenvector centrality in Eqn.(6) and simply use $\alpha$\texttt{=}$0.5$ to represent general Katz centrality in Eqn.(8).

\textbf{Ablation on scaling parameters $\gamma$ and $\beta$.} Table S\ref{tab:ab_gamma} shows MCL results with different scaling parameters $\gamma$ and $\beta$. As discussed in Sec.5 that a larger scaling parameter (\ie, a smaller temperature in softmax-like random walk probability) will lead to a more concentrated eigenvector centrality. However, a larger scaling parameter would inevitably bias the meta-training. Thus, there exists a trade-off to select optimal parameters according to each task.

In the experiments, we find its empirically effective to select $\gamma$ and $\beta$ according to their pre-trained models (similar to Table S\ref{tab:ab_gamma}). In most cases, we use $\gamma$\texttt{=}$20$, $\beta$\texttt{=}$10$ and $\gamma$\texttt{=}$40$, $\beta$\texttt{=}$20$ for 1-shot and 5-shot tasks, respectively.  

\section{Additional Plugin Experiments}

We have shown that our proposed centrality weighted pooling has a consistent performance gain (especially in the extreme 1-shot scenario) over global average pooling on ProtoNet \cite{snell2017prototypical} and RelationNet \cite{sung2018learning} by concentrating on more task-relevant local features. Besides Table 1, 2, 4 and S\ref{tab:wometa}, we give additional results in Table S\ref{tab:add_plugin} to show that MCL can be easily plugged into those methods that need no extra parameters except for the feature extractor.

The experiments are conducted by the following rules: all comparing methods are evaluated with the same pre-trained backbone ResNet-12 as in Sec.\ref{sec:wometa} without meta-training; we only use centrality weighted pooling to aggregate local features from query image as different methods have different operations on support features; we fixed the parameters $\gamma$\texttt{=}$20$ and $\beta$\texttt{=}$10$ to MCL plugins for all comparing methods.

As shown in Table S\ref{tab:add_plugin}, our proposed centrality plugin consistently improves the performances of all the global feature based methods without bells and whistles.

\begin{table}[t]
\newcommand{\CC}{\cellcolor{Gray}}
\centering
\small
\aboverulesep=0ex
\belowrulesep=0.5ex
\setlength{\tabcolsep}{3pt}
\resizebox{\linewidth}{!}{%
\begin{tabular}{lllll}
\toprule
\multirow{2}{*}{\textbf{Method}} & \multicolumn{2}{l}{\textbf{\textit{mini}ImageNet}} & \multicolumn{2}{l}{\textbf{\textit{tiered}ImageNet}} \\
 & 1-shot & 5-shot & 1-shot & 5-shot \\
\midrule
Baseline\cite{chen2019closer} & 53.99 & 78.78 & 67.75 & 85.23 \\
\CC +MCL & \CC 56.21\textcolor{red}{\scriptsize{+2.22}} & \CC 80.44\textcolor{red}{\scriptsize{+1.66}} & \CC 68.53\textcolor{red}{\scriptsize{+0.78}} & \CC 85.75\textcolor{red}{\scriptsize{+0.52}} \\
\midrule
Baseline++\cite{chen2019closer} & 55.03 & 78.79 & 61.86 & 84.31 \\
\CC +MCL & \CC 56.93\textcolor{red}{\scriptsize{+1.90}} & \CC 79.93\textcolor{red}{\scriptsize{+1.14}} & \CC 63.00\textcolor{red}{\scriptsize{+1.14}} & \CC 84.61\textcolor{red}{\scriptsize{+0.30}} \\
 \midrule
ProtoNet\cite{snell2017prototypical} & 56.50 & 79.68 & 64.27 & 84.01 \\
\CC +MCL & \CC 58.62\textcolor{red}{\scriptsize{+2.12}} & \CC 80.99\textcolor{red}{\scriptsize{+1.30}} & \CC 66.48\textcolor{red}{\scriptsize{+2.21}} & \CC 84.83\textcolor{red}{\scriptsize{+0.82}} \\
\midrule
MatchingNet\cite{vinyals2016matching}  & 58.41 & 79.52 & 68.68 & 85.16 \\
\CC +MCL & \CC 59.85\textcolor{red}{\scriptsize{+1.44}} & \CC 80.79\textcolor{red}{\scriptsize{+1.27}} & \CC 69.45\textcolor{red}{\scriptsize{+0.97}} & \CC 85.49\textcolor{red}{\scriptsize{+0.33}}\\
 \midrule
R2D2\cite{bertinetto2018meta}  & 59.82 & 78.97 & 70.22 & 85.30 \\
\CC +MCL & \CC 60.86\textcolor{red}{\scriptsize{+1.04}} & \CC 80.65\textcolor{red}{\scriptsize{+1.68}} & \CC 70.64 \textcolor{red}{\scriptsize{+0.42}} & \CC 85.84\textcolor{red}{\scriptsize{+0.54}}\\
 \midrule
MetaOptNet\cite{lee2019meta}  & 59.59 & 79.77 & 69.55 & 85.25 \\
\CC +MCL & \CC 60.66\textcolor{red}{\scriptsize{+1.07}} & \CC 81.27\textcolor{red}{\scriptsize{+1.50}} & \CC 70.11\textcolor{red}{\scriptsize{+0.55}} & \CC 85.76\textcolor{red}{\scriptsize{+0.51}}\\
 \midrule
DSN\cite{simon2020adaptive} &  58.70 & 79.01 & 69.13 & 85.14 \\
\CC +MCL & \CC 60.08\textcolor{red}{\scriptsize{+1.38}} & \CC 80.66\textcolor{red}{\scriptsize{+1.65}} & \CC 69.59\textcolor{red}{\scriptsize{+0.46}} & \CC 85.61\textcolor{red}{\scriptsize{+0.47}}\\
\midrule
Neg-cosine\cite{liu2020negative} &  58.82 & 79.63 & 69.44 & 84.94 \\
\CC +MCL & \CC 60.20\textcolor{red}{\scriptsize{+1.38}} & \CC 81.04\textcolor{red}{\scriptsize{+1.41}} & \CC 69.85\textcolor{red}{\scriptsize{+0.41}} & \CC 85.31\textcolor{red}{\scriptsize{+0.37}}\\
 \bottomrule
\end{tabular}
}
\caption{Few-shot classifications (\%) before and after applying centrality weighted pooling. Experiment settings follow Table S\ref{tab:wometa}. \label{tab:add_plugin}}
\end{table}

\section{Additional Visualization}

The additional Grad-CAM visualizations as in Table 5 of the main paper are presented as follows: Table S\ref{tab:e2e} illustrates end-to-end MCL-Katz like in Table 5(b). Table S\ref{tab:proto_plugin} illustrates MCL plugin on ProtoNet \cite{snell2017prototypical} like in Table 5(a).

\begin{table*}[t]
\centering
\begin{tabular}{cc}
\includegraphics[width=0.48\textwidth]{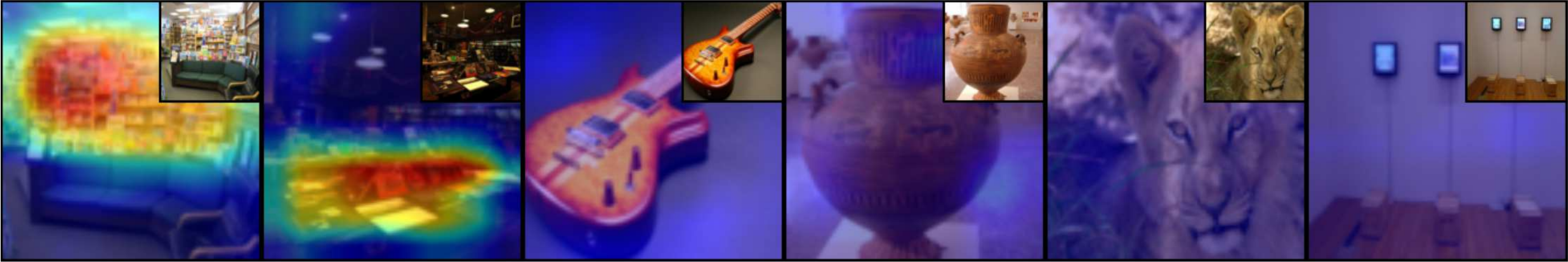} &
\includegraphics[width=0.48\textwidth]{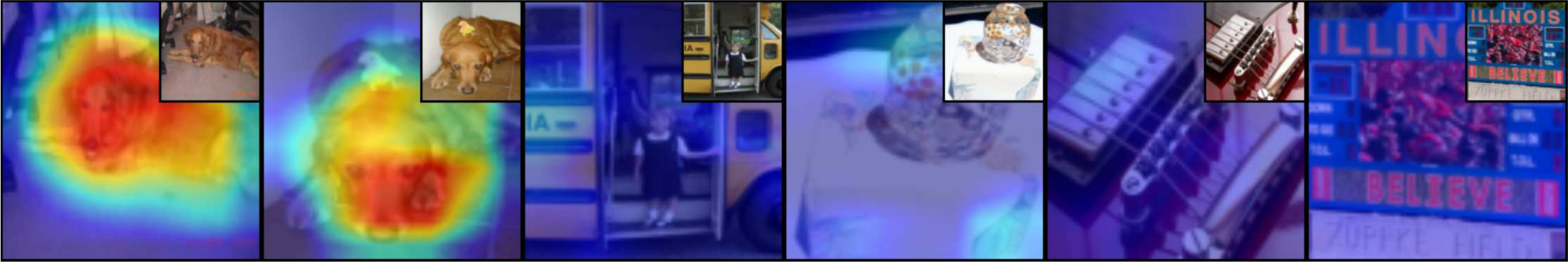} \\
\includegraphics[width=0.48\textwidth]{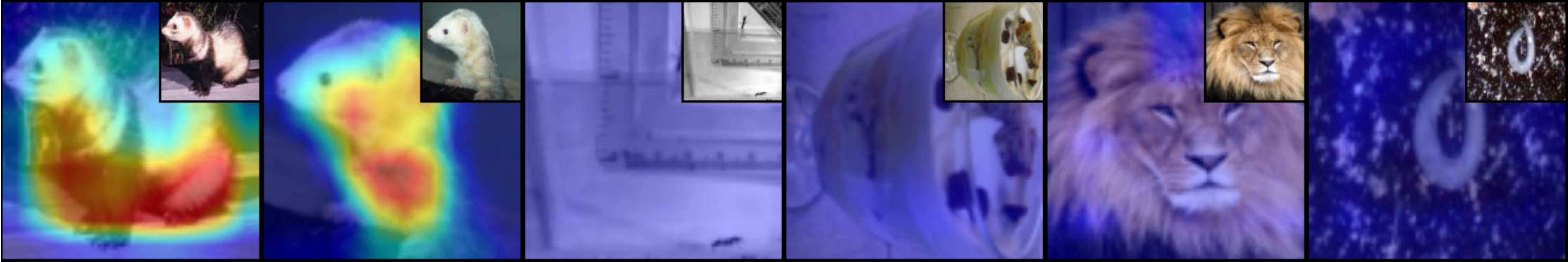} &
\includegraphics[width=0.48\textwidth]{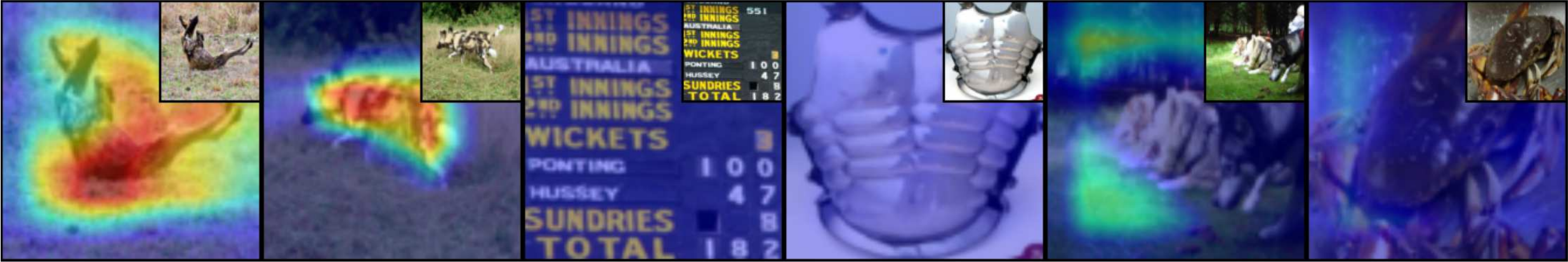} \\
\includegraphics[width=0.48\textwidth]{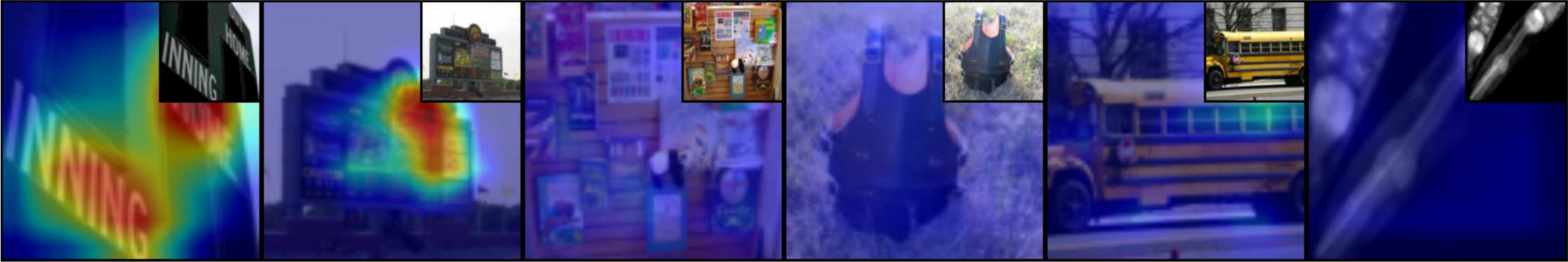} &
\includegraphics[width=0.48\textwidth]{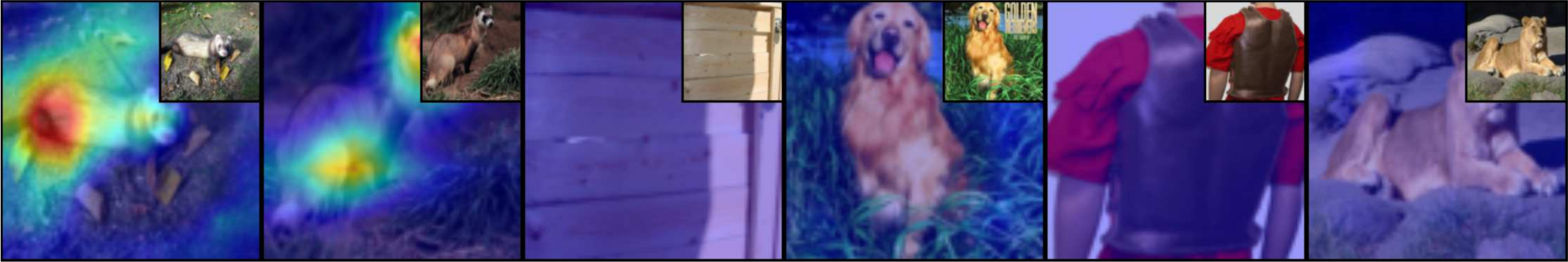} \\
\includegraphics[width=0.48\textwidth]{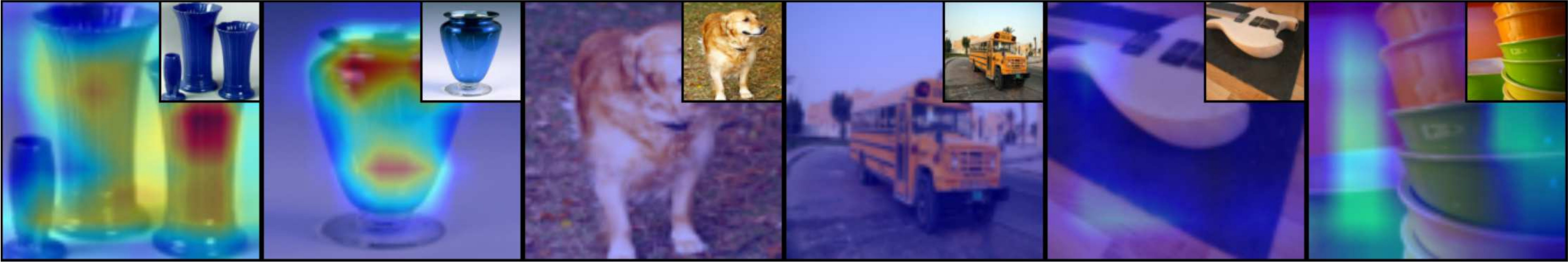} &
\includegraphics[width=0.48\textwidth]{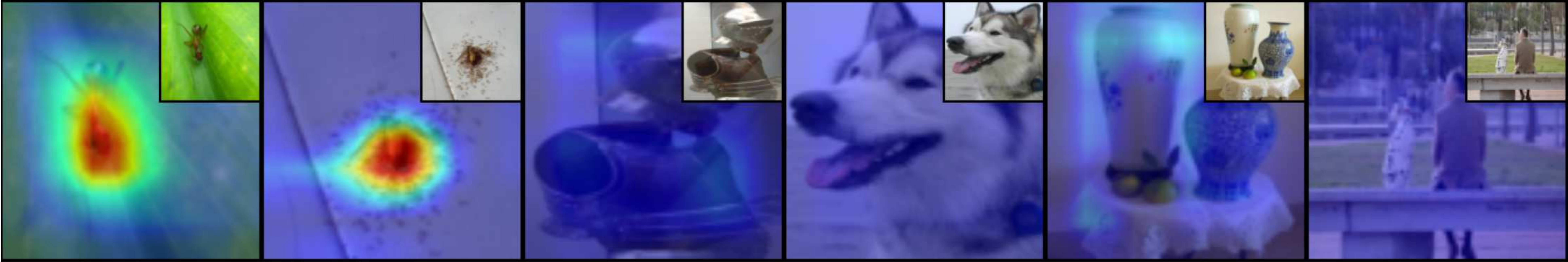} \\
\includegraphics[width=0.48\textwidth]{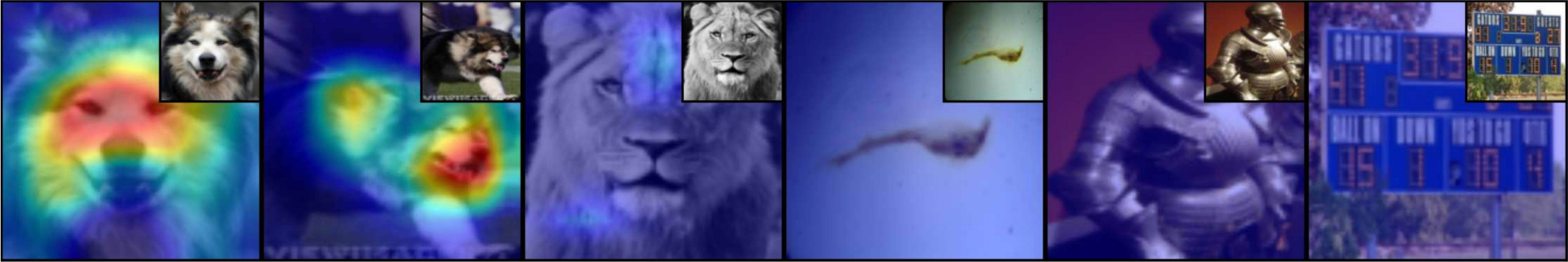} &
\includegraphics[width=0.48\textwidth]{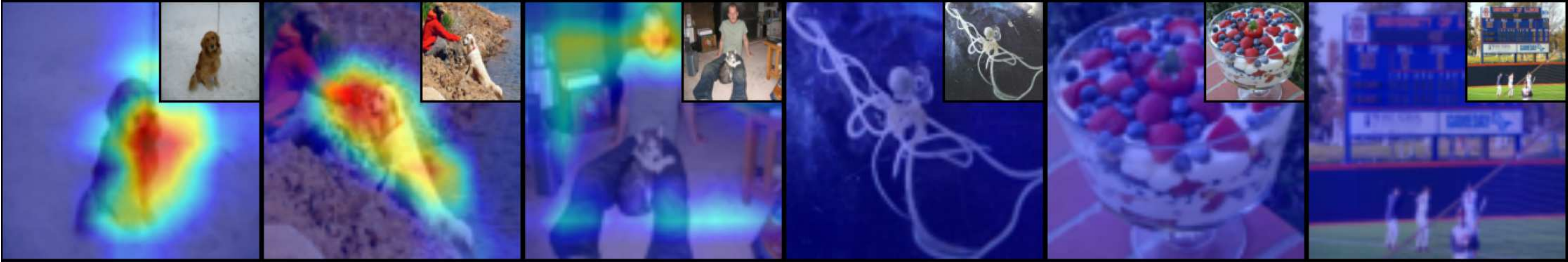} \\
\includegraphics[width=0.48\textwidth]{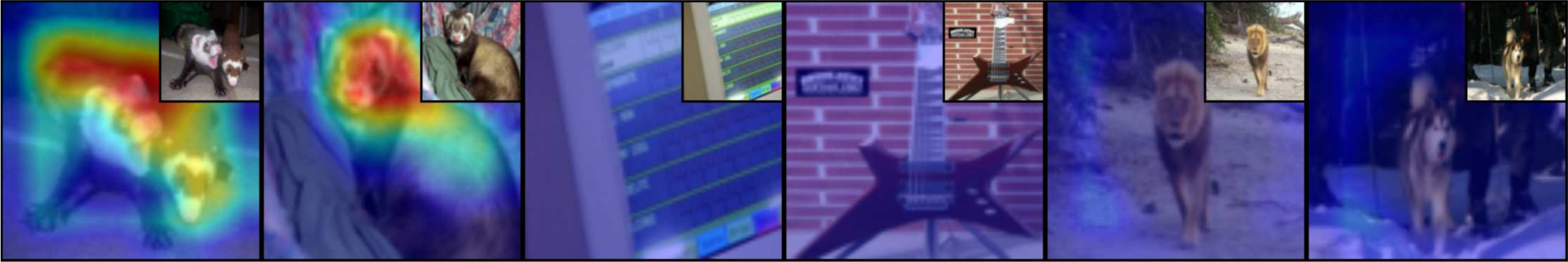} &
\includegraphics[width=0.48\textwidth]{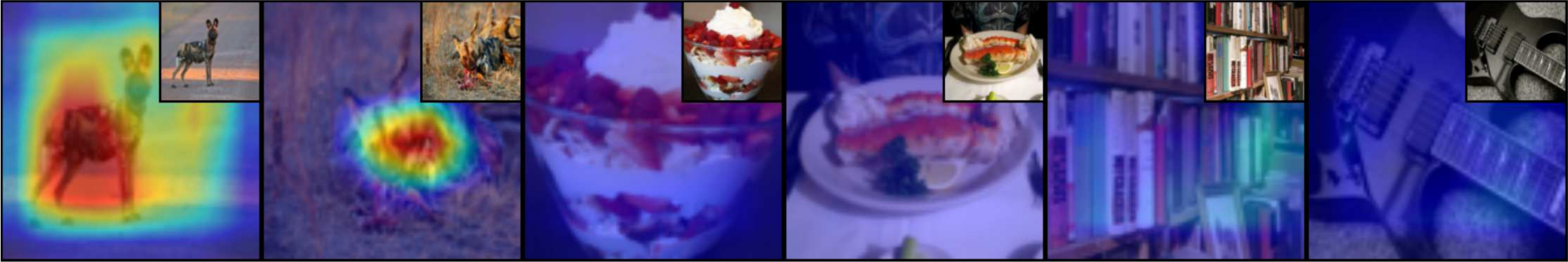} \\
\includegraphics[width=0.48\textwidth]{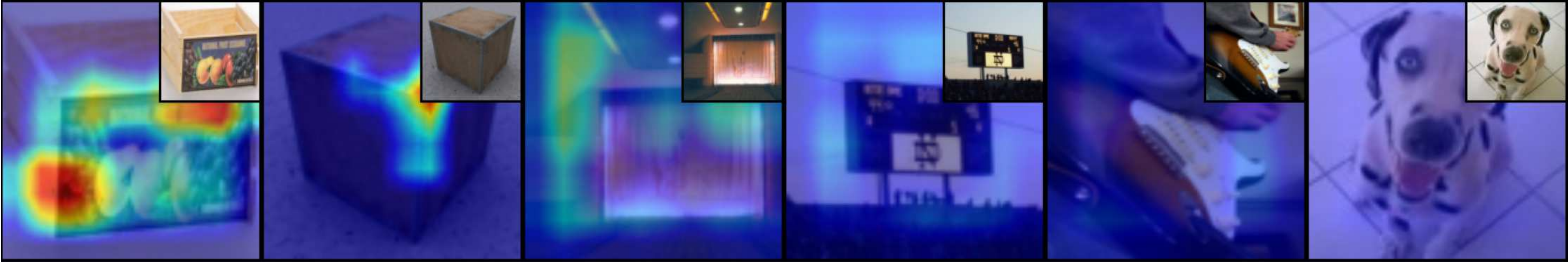} &
\includegraphics[width=0.48\textwidth]{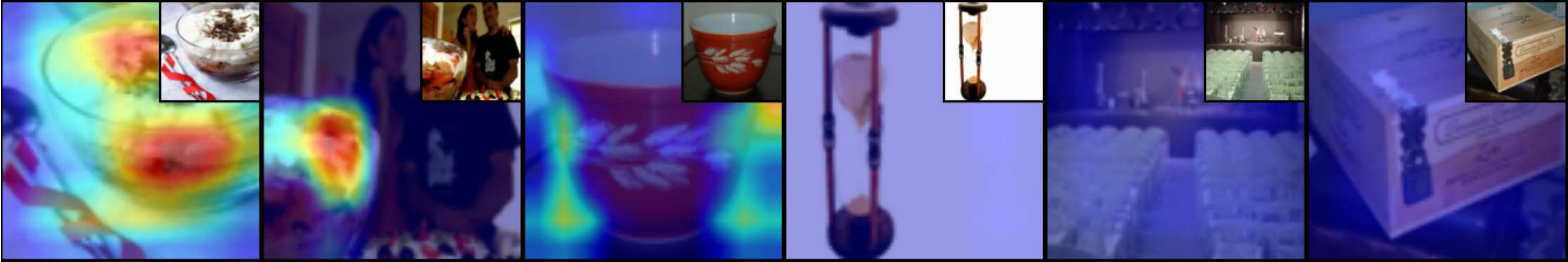} \\
\end{tabular}
\caption{Additional Grad-CAM visualizations of MCL-katz on 5-way 1-shot FSL tasks (\textit{mini}ImageNet) with ResNet-12. Formatting follows Table 5(b): query images are placed in the first column for each task; ground truth support images are placed in the second column; the four images on the far right column of each task are from the confounding support classes.\label{tab:e2e}}
\end{table*}

\clearpage

\begin{table*}[t]
\centering
\begin{tabular}{cc}
 \includegraphics[width=0.45\textwidth]{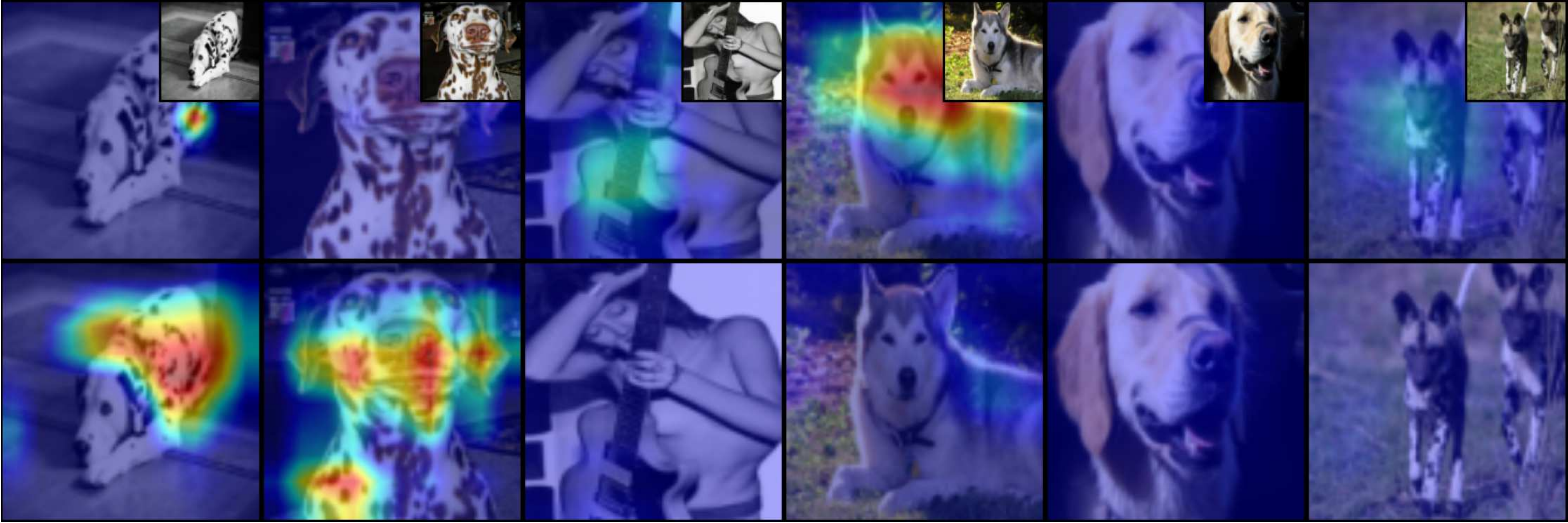} &
\includegraphics[width=0.45\textwidth]{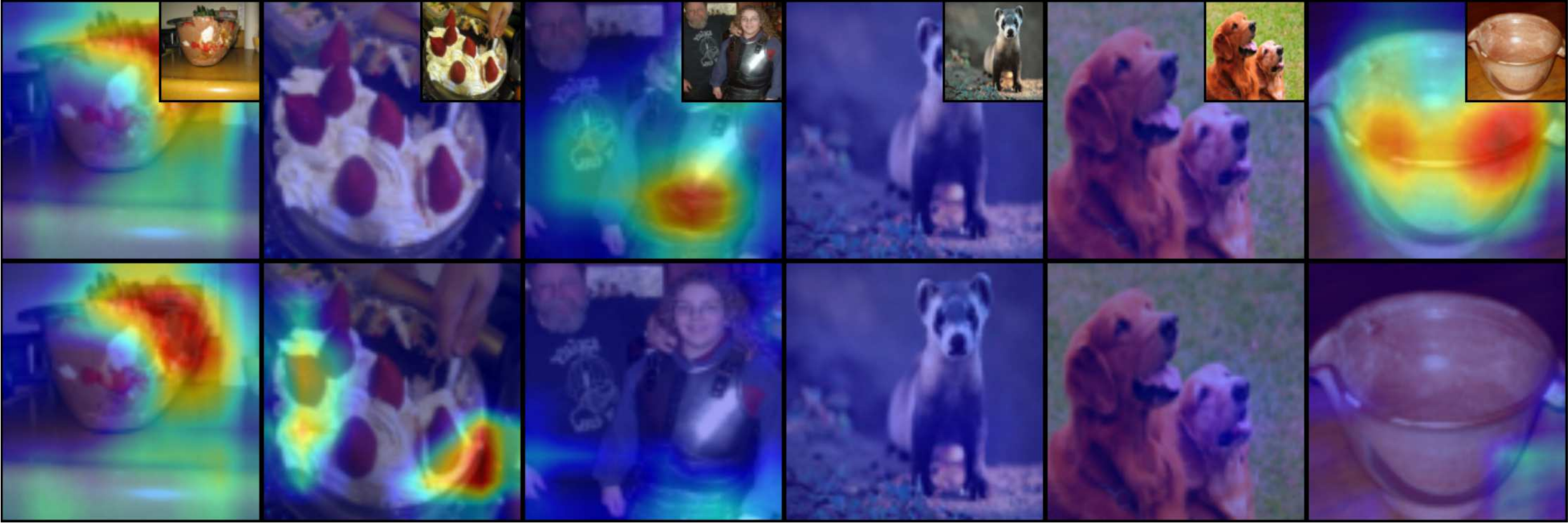}
\\
 \includegraphics[width=0.45\textwidth]{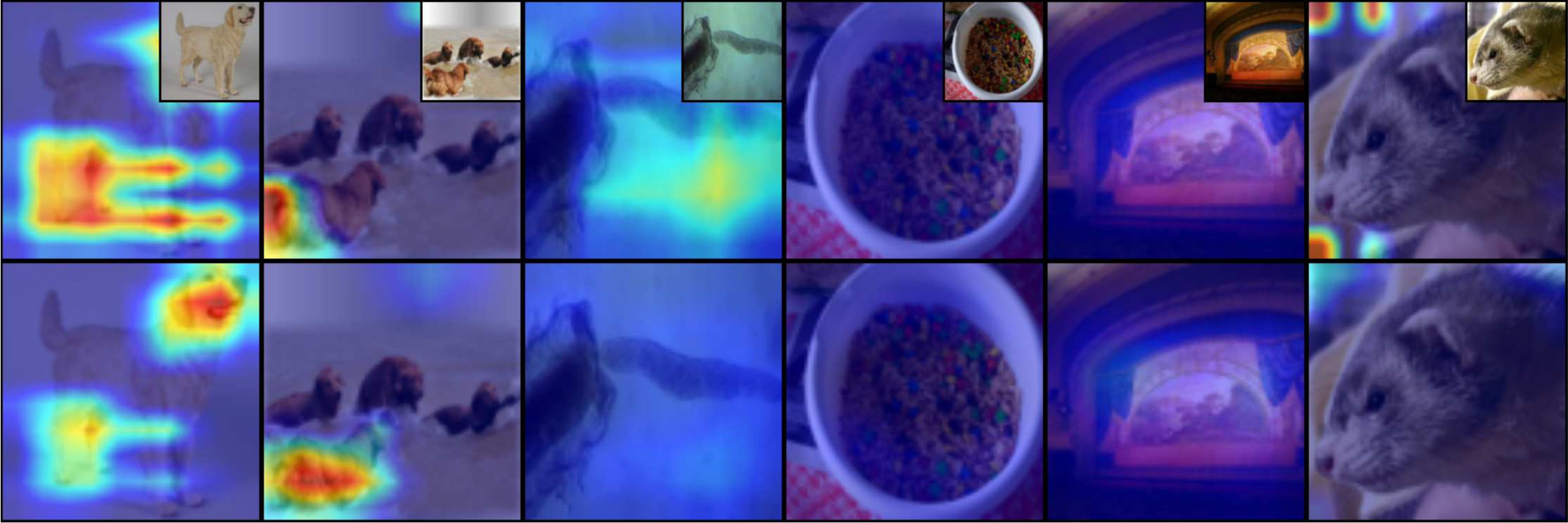} &
\includegraphics[width=0.45\textwidth]{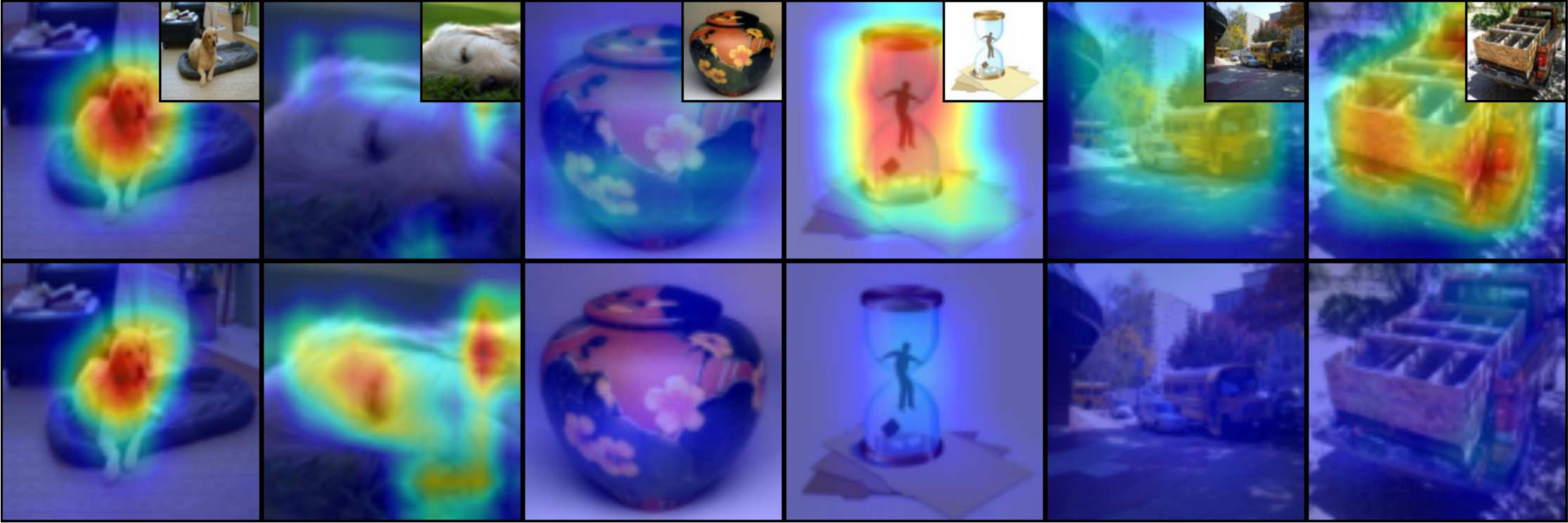}
\\
\includegraphics[width=0.45\textwidth]{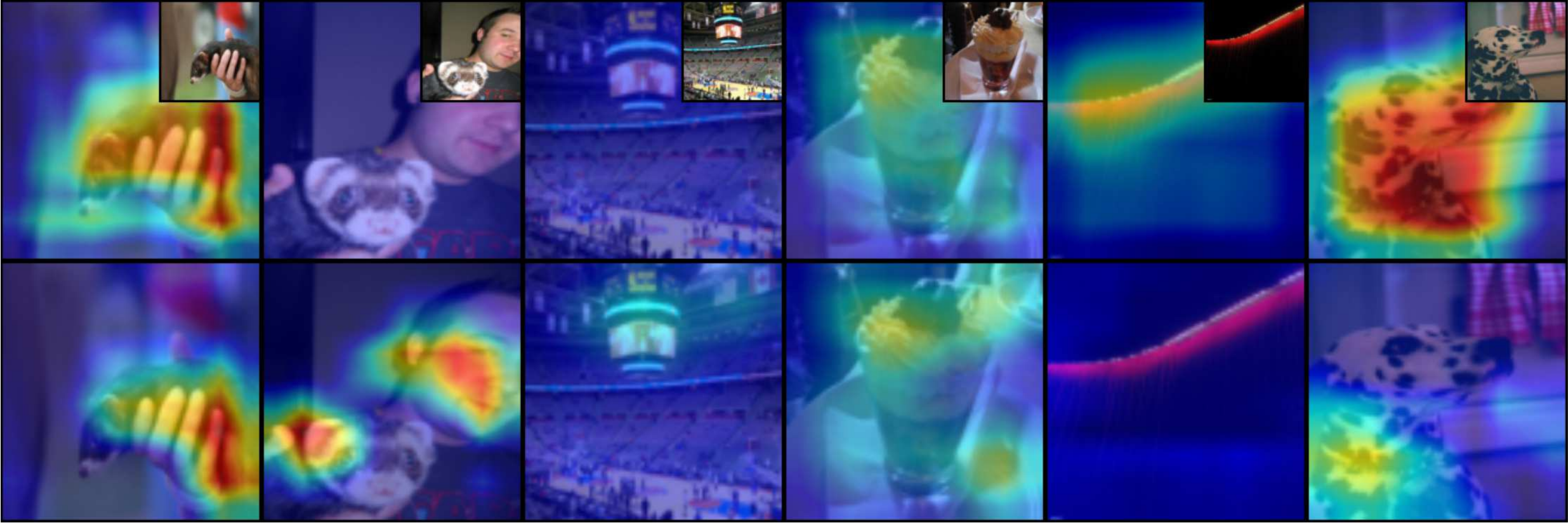} &
\includegraphics[width=0.45\textwidth]{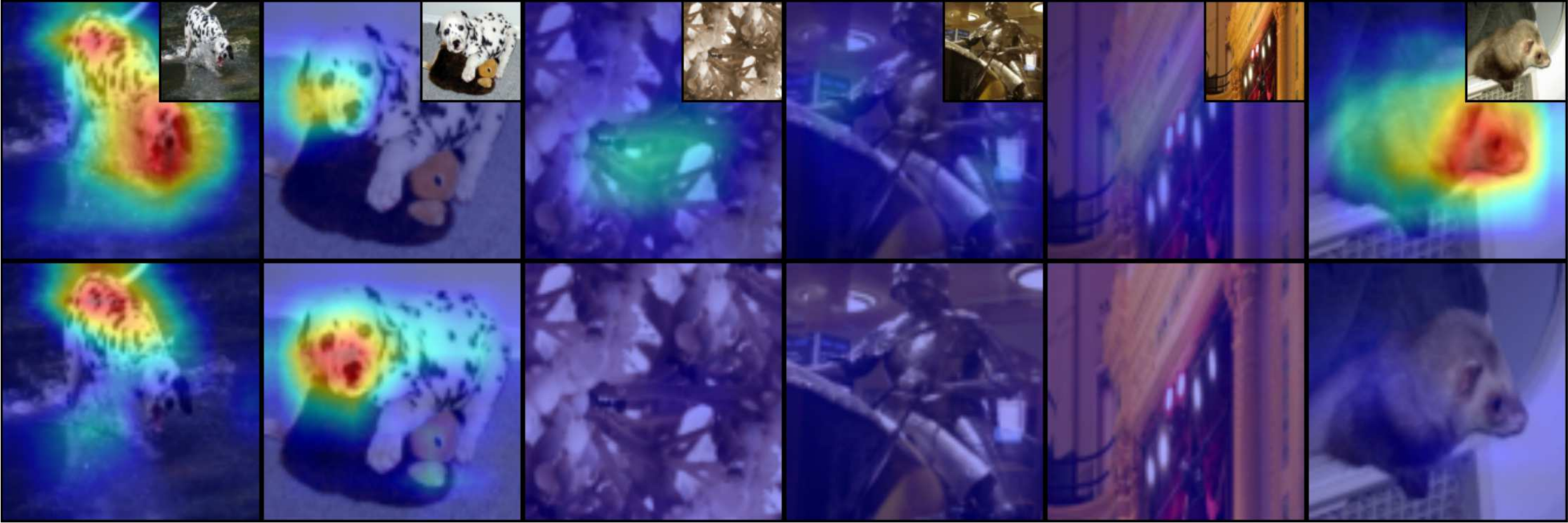}
\\
 \includegraphics[width=0.45\textwidth]{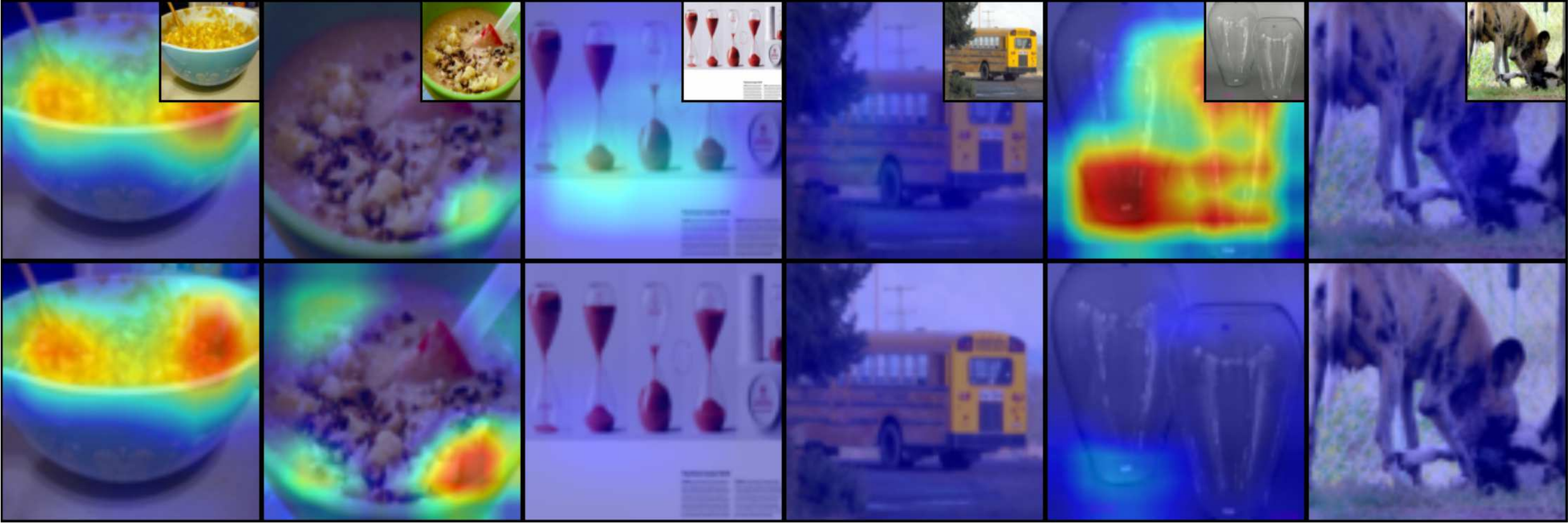} &
\includegraphics[width=0.45\textwidth]{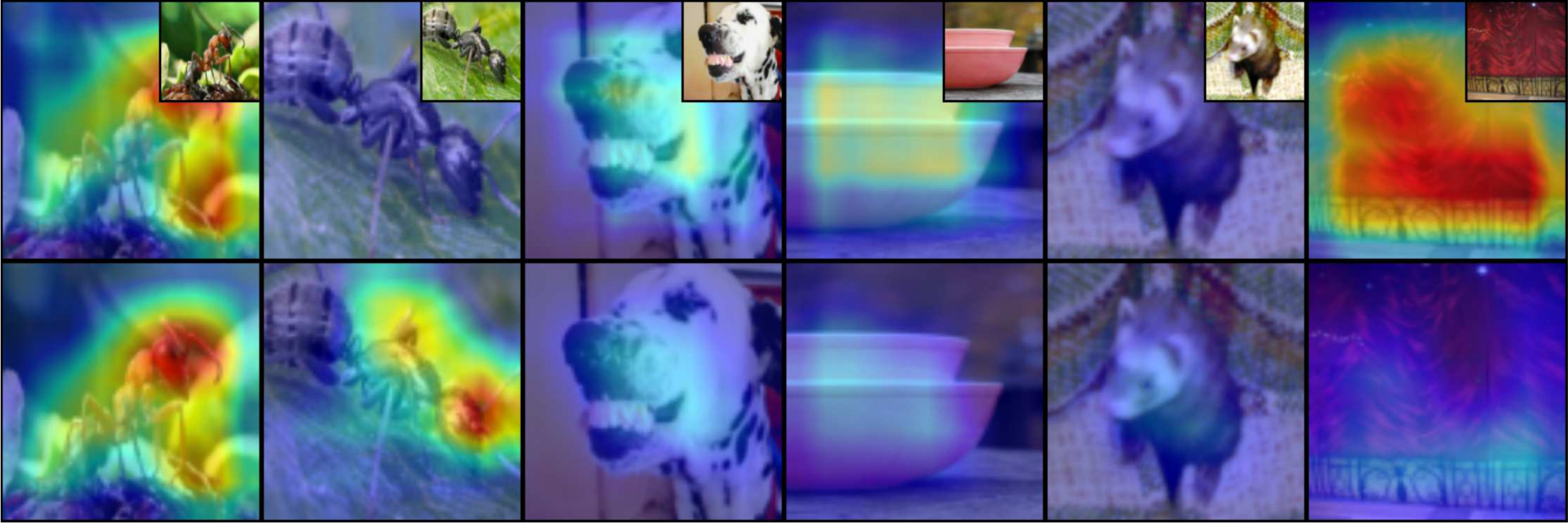}
\\
 \includegraphics[width=0.45\textwidth]{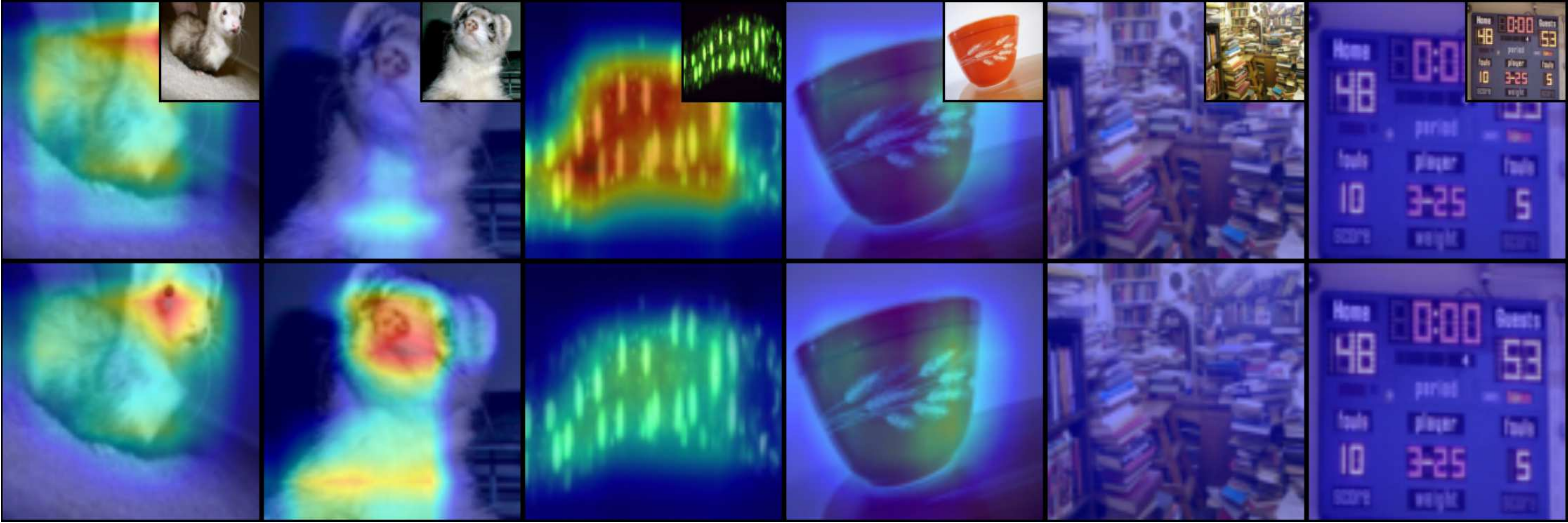} &
\includegraphics[width=0.45\textwidth]{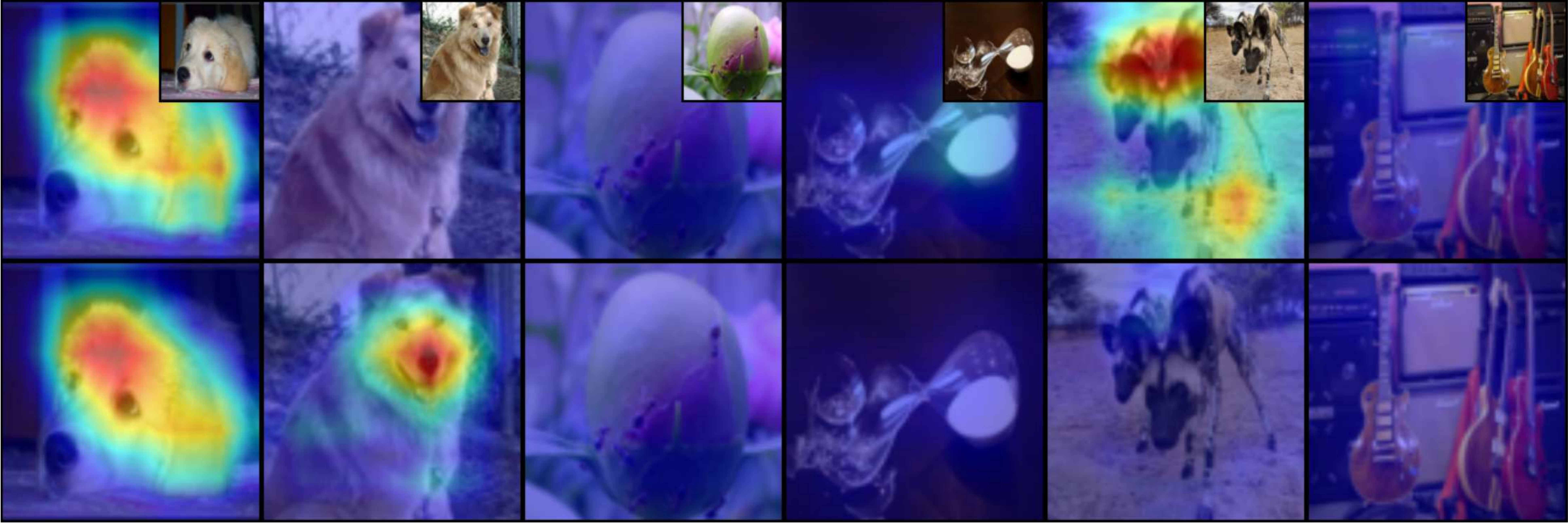}
\\
 \includegraphics[width=0.45\textwidth]{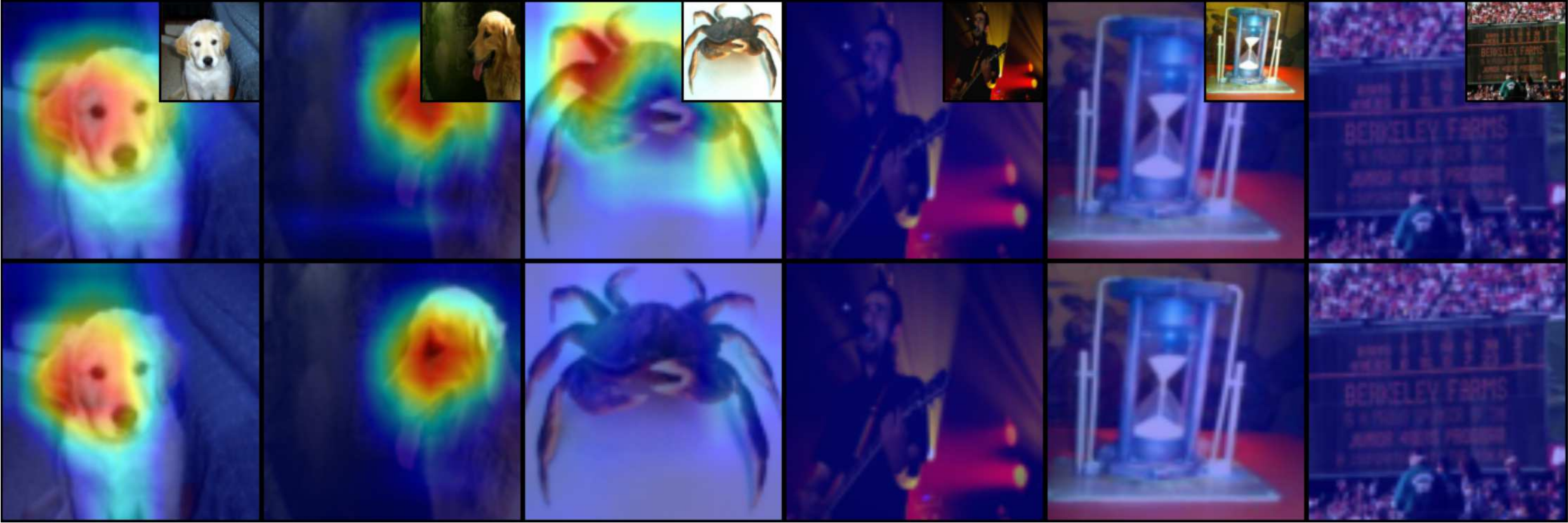} &
\includegraphics[width=0.45\textwidth]{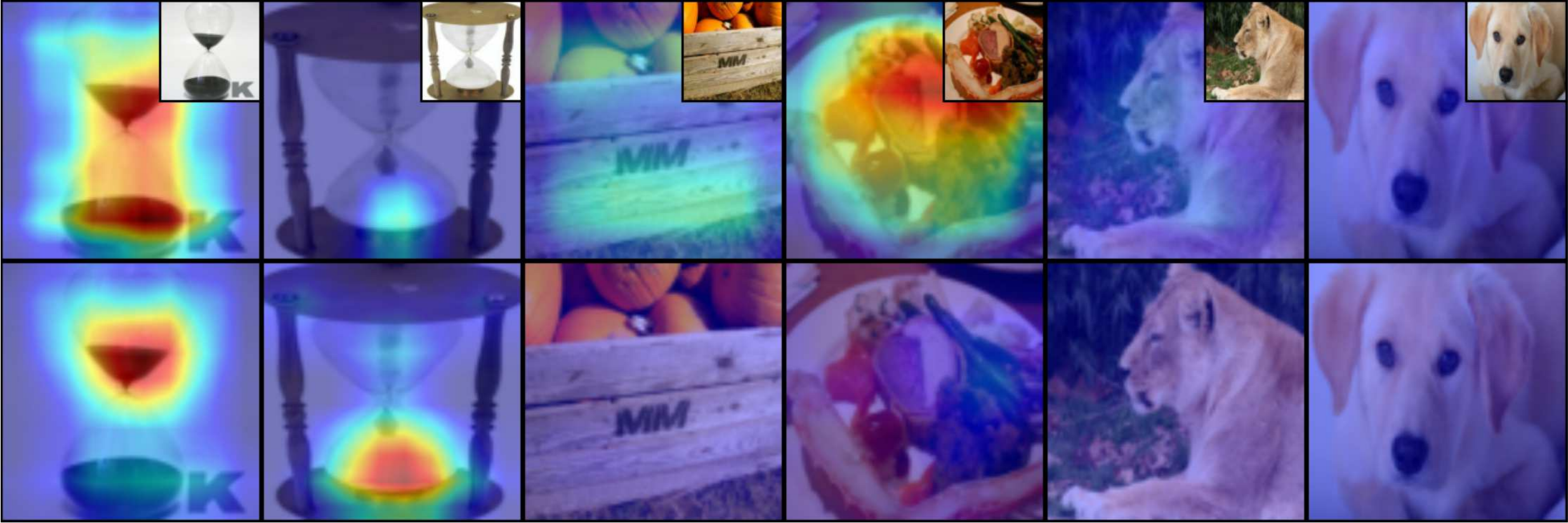}
\\
 \includegraphics[width=0.45\textwidth]{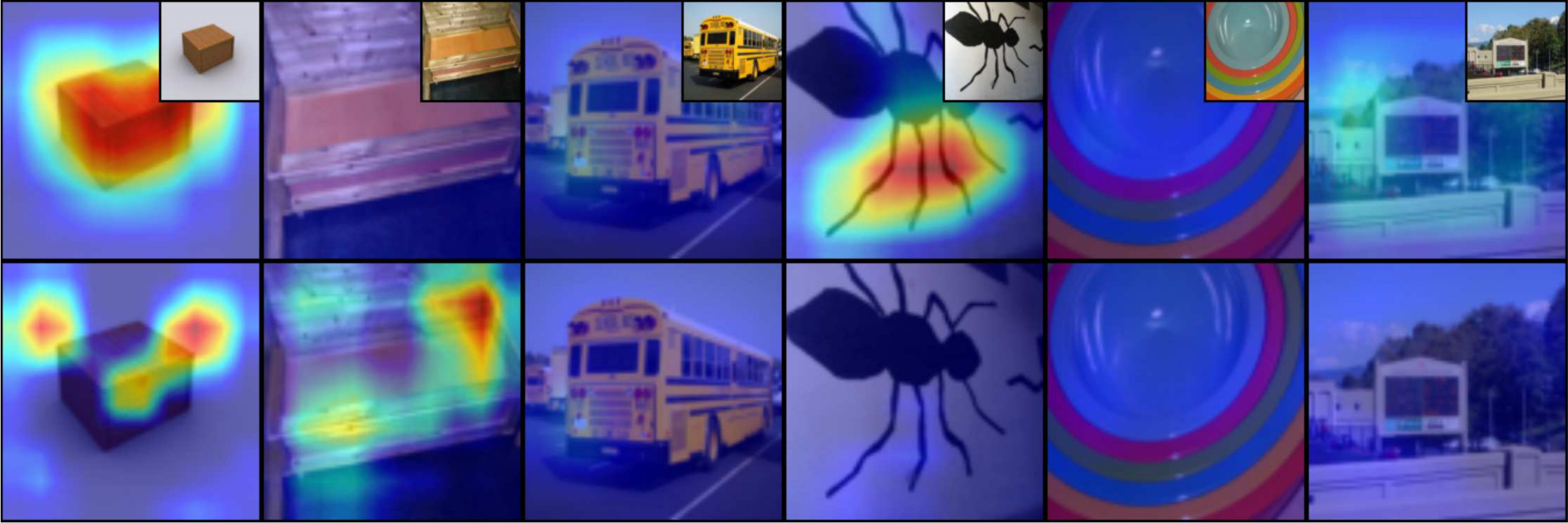} &
\includegraphics[width=0.45\textwidth]{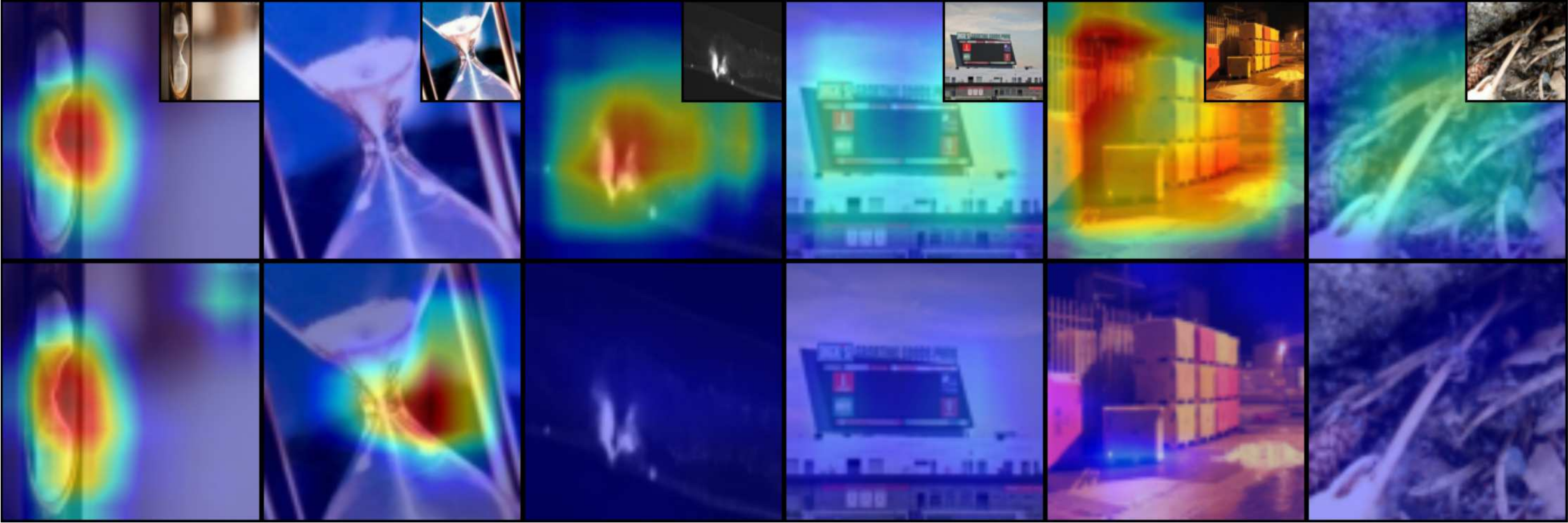}
\end{tabular}
\caption{Additional \textbf{ProtoNet+MCL} Grad-CAM visualizations on 5-way 1-shot FSL tasks (\textit{mini}ImageNet) with ResNet-12. Formatting follows Table 5(a): query images are placed in the first column for each task; ground truth support images are placed in the second column; the four images on the far right column of each task are from the confounding support classes. The top row in each task is from ProtoNet while the bottom row is from ProtoNet+MCL.\label{tab:proto_plugin}}
\end{table*}

\end{document}